\documentclass[%
  onecolumn
]{mpi2015-cscpreprint}

\usepackage[american]{babel}

\usepackage{graphicx}

\usepackage{amssymb}
\usepackage{amsthm}
\usepackage[american]{babel}

\usepackage{amssymb,amsfonts,amsmath}
\usepackage{subcaption}

\usepackage{url}
\usepackage{float} 
\usepackage{amssymb,amsfonts,amsmath}
\usepackage{graphicx,color,changebar}
\usepackage{algorithm}
\usepackage{multirow}
\usepackage[noend]{algpseudocode}
\usepackage[utf8]{inputenc}
\usepackage{circuitikz}
\usepackage{comment}
\usepackage{chemmacros}
\usepackage{cleveref}
\usepackage{hyperref}
\usepackage{siunitx}
\usepackage{esvect}

\begin{document}
  

\title{Data-Augmented Predictive Deep Neural Network: Enhancing the extrapolation capabilities of non-intrusive surrogate models}

\author[$\ast$]{Shuwen Sun}
\affil[$\ast$]{Max Planck Institute for Dynamics of Complex Technical Systems\authorcr
  \email{ssun@mpi-magdeburg.mpg.de;}}
\author[$\ast$]{Lihong Feng}
\author[$\ast$]{Peter Benner}
\shorttitle{Data-Augmented Predictive Deep Neural Network}
\shortauthor{Sun, Feng and etc.}
\shortdate{}
  

\abstract{%
Numerically solving a large parametric nonlinear dynamical system is challenging due to its high complexity and the high computational costs. In recent years, machine-learning-aided surrogates are being actively researched. However, many methods fail in accurately generalizing in the entire time interval $[0, T]$, when the training data is available only in a training time interval $[0, T_0]$, with $T_0<T$.

To improve the extrapolation capabilities of the surrogate models in the entire time domain, we propose a new deep learning framework, where kernel dynamic mode decomposition (KDMD) is employed to evolve the dynamics of the latent space generated by the encoder part of a convolutional autoencoder (CAE). After adding the KDMD-decoder-extrapolated data into the original data set, we train the CAE along with a feed-forward deep neural network using the augmented data. The trained network can predict future states outside the training time interval at any out-of-training parameter samples. The proposed method is tested on two numerical examples: a FitzHugh-Nagumo model and a model of incompressible flow past a cylinder. Numerical results show accurate and fast prediction performance in both the time and the parameter domain.
}

\novelty{Using only the training data in the time interval $[0, T_0]$, the proposed Data-Augmented Predictive Deep Neural Network (DAPredDNN) makes use of augmented data generated by KDMD to realize time extrapolation in the time interval $[T_0, T]$, $T_0 < T$. Unlike some existing methods that rely on sequential predictions, DAPredDNN completes the predictions in a single step.}

\maketitle


\section{Introduction}%
\label{sec:intro}

Efficiently predicting dynamics of a high-fidelity model in a multi-query task is challenging but of high interest in many research fields and applications. Direct simulation usually requires enormous computing power and needs long computation time. A possible way to solve such issues is to construct a reduced-order model (ROM) with a compact size at the offline stage. At the online stage, this ROM can act as a surrogate for the original large model to achieve fast and cheap predictions.

Traditional intrusive model order reduction (MOR) methods based on projection \cite{morGugA04, morBenGW15, morhandbookV1, morhandbookV2, morhandbookV3, morChaS10, morQuaR14} require accessibility of the numerically discretized operators, making them difficult to be applied to problems whose discretization information for the projection-based MOR is hard to be extracted.

Motivated by the demanding needs for MOR of such problems, non-intrusive MOR methods are being actively studied, among which machine learning-aided MOR methods have been proposed \cite{morFreDM21, morFreM22, morNikKP22, morBruK22, morLeeC20}, where an autoencoder (AE) is usually used for dimension reduction. However, many of them fail in extrapolation in the time domain. The method in \cite{morMauLB21} considering CAE and long short-term memory (LSTM) gives good prediction in the parameter space within the training time interval $[0, T_0]$ for parametric fluid problems. However, extrapolation of the solution in time, i.e., prediction of the solution at any time instant $t, t > T_0$, is not considered. At the online stage, a high-fidelity solution in a time window corresponding to any new parameter sample is still needed. This entails extra computations depending on the high-fidelity model simulations at the online prediction phase. The method in \cite{morFreFM23} utilizes LSTM encoder-decoder to realize future dynamics prediction. It can achieve a better prediction at future time steps outside the time interval of the training data. The LSTM in these work acts as an iterative predictor, processing $p$-step to $k$-step prediction. This means that the predicted dynamics at the previous $p$ time instants need to be fed into the LSTM to predict the dynamics at the future $k$ time instants. A method using temporal convolutional autoencoder is proposed in \cite{XuK20} to predict future states. However, the prediction also depends on the precomputed high-fidelity model solutions in a time window, similar as the LSTM approach in \cite{morMauLB21}. Moreover, the future dynamics can only be predicted step-by-step, leading to a slow online prediction.

Another approach to predicting future dynamics is to learn a system of ordinary differential equations (ODEs) from time series data and then use the learned ODE to predict the state at future time steps. Sparse identification of nonlinear dynamics (SINDy) \cite{morBruPK16} learns the governing equations of a dynamical system by identifying sparse combinations of functions in a library. SINDy is efficient only for systems of small dimensions. Recently, SINDy has been combined with AE in \cite{morChaLKetal19} to realize prediction for large-scale systems. In \cite{morConGFetal23}, Paolo et al. develop a framework where proper orthogonal decomposition and AE are applied for dimension reduction while a parametric SINDy structure is responsible for identification of the latent space dynamics. Similar work \cite{FrHC22} named as Latent Space Dynamics Identification (LaSDI) can also be fitted within the AE-SINDy framework. The framework is parametrized by interpolating the coefficients of SINDy-selected functions corresponding to training parameters. Further details of LaSDI and its variants are reviewed in \cite{BonHTetal24}. These methods achieve good predictions in both the parameter and the time domain via time-marching schemes in the latent space. Both LSTM-based methods~\cite{morMauLB21, morFreFM23} and AE-SINDy based methods~\cite{FrHC22, morChaLKetal19, morConGFetal23} can be seen as the ($p$)-step to ($k$)-step prediction at the online phase. Such prediction processes depend on the previous $p$-time steps to predict the future $k$ time steps. Transformers are introduced in \cite{morGenZ22, morSolVGetal24} to mimic the dynamics in the latent space , However, the method introduced in \cite{morSolVGetal24} is primarily based on non-parametric problems. In \cite{morGenZ22}, the proposed method's ability to predict the future solution in the time interval $[T_0, T]$ without any training data available in that interval, remains unclear.

Recently, dynamic mode decomposition (DMD) and its variants have been connected to deep neural networks to explore different dimension-reduction processes. Inspired by the extended DMD (EDMD) method, Li et al. \cite{morLiDBetal17} make use of an AE to learn the optimal observations for EDMD so that a more accurate Koopman operator can be given. Compared to this dictionary-searching architecture, Otto et al. in \cite{OttR19} use linearly recurrent autoencoder networks (LRAN) to seek a Koopman invariant subspace of observables. Lusch et al. \cite{LusKB18} focus on the Koopman theory and combine it with AE to find an interpretable representation of the Koopman eigenfunctions. These autoencoder-generated eigenfunctions represent a reduced coordinate system in the latent space. All these three methods mentioned above haven not included any parameter - latent variable mapping, and are not yet applicable to the parametric cases. Parametric DMD with the help of CAE is proposed in \cite{DuaH24}. The parametric dependencies are realized by interpolation of the latent variable. However, the accuracy is not satisfying especially at the time-extrapolation phase.

In this work, we propose a new framework named Data-Augmented Predictive Deep Neural Network (DAPredDNN) that combines Kernel Dynamic Mode Decomposition (KDMD) \cite{morWilRK15} with DNNs, aiming to predict the system dynamics in the parameter-time domain. Specifically, our approach yields accurate predictions in $[0, T]$ for test parameter samples based on the training data collected at training parameter samples in the time interval $[0, T_0]$, where $T_0 < T$. The proposed method employs CAE for compressing the original high-dimensional parametric data into a latent space with a very small dimension. The FFNN maps parameter-time pairs to this latent space. During the training stage, KDMD derives the latent variables in the interval $(T_0, T]$ based on the latent data from $[0, T_0]$ obtained via the pretrained encoder. The decoder is used to recover the physical dynamics from the latent variables in $(T_0, T]$. Those data in $(T_0, T]$ are combined with the original training data in $[0, T_0]$ to train CAE-FFNN without regenerating any new data in $(T_0, T]$ by simulating the high-fidelity model. The trained FFNN-decoder builds a direct mapping between the parameter-time domain and the corresponding dynamics in the entire time interval $[0, T]$. During the online prediction, FFNN-decoder can generate set of sequences of future dynamics corresponding to any set of testing parameters in only one step. In contrast to the existing $p$-step to $k$-step or step-by-step predictions, this one-step time sequence prediction makes the online prediction extremely fast.

The rest of this paper is outlined as follows. In \Cref{sec:backgrd}, a brief overview of AE and DMD is provided. The main part of the new network architecture is introduced in \Cref{sec:work}. In this part, some basic building blocks of the proposed framework are displayed in \Cref{subsec:blocks}. The comprehensive architecture of the DAPredDNN and the detailed algorithms are presented in \Cref{subsec:aug_data}. The proposed framework is tested and validated with two numerical examples including a FitzHugh–Nagumo model and the flow past a cylinder in \Cref{sec:examples}. \Cref{sec:conclu}  concludes the paper and suggests further investigations.

\section{Background}%
\label{sec:backgrd}
 
\subsection{Problem setting}
\label{subsec:problem}
We consider dynamics of problems described by nonlinear time-dependent parametric partial differential equations (PDEs). After numerical discretization in the space domain, the resulting ODEs can be written as:
\begin{equation}
\label{eq:ode}
\left\{\begin{array}{l}
\frac{d}{dt} \boldsymbol{u}_h(\boldsymbol{\mu}, t) = \boldsymbol{f}\left(\boldsymbol{u}_h(\boldsymbol{\mu}, t), \boldsymbol{\mu}, t \right), \\
\boldsymbol{u}_h(\boldsymbol{\mu}, t=0) = \boldsymbol{u}_1(\boldsymbol{\mu}),
\end{array}\right.
\end{equation}
where $\boldsymbol{u}_h \in \mathbb{R}^N$ is the full state variable. The vector $\boldsymbol{\mu} \in \mathbb{R}^{d_p}$ contains $d_p$ parameters of interest and the time $t \in [0, T]$. $\boldsymbol{f}: \mathbb{R}^N \times \mathbb{R}^{p} \times [0, T] \rightarrow \mathbb{R}^{N}$ is usually a nonlinear function of $\boldsymbol{u}_h$, $\boldsymbol{\mu}$ and $t$, describing the nonlinear dynamics of the system. The discretized system, also referred to as the full order model (FOM), usually has a large spatial dimension, denoted by $N$. This work aims to propose a non-intrusive method that establishes NN-based surrogates of this dynamical system, such that the parametric behaviour and time evolution of the system dynamics can be quickly obtained. To train the framework, we use numerically simulated solution data from, e.g., black-box software. The data is assembled into:
\begin{equation}
\label{eq:snapshots}
\boldsymbol{U}_h(\boldsymbol{\mu}_i)=\left[\begin{array}{cccc}
\mid & \mid & & \mid \\
\boldsymbol{u}_{h}(\boldsymbol{\mu}_i, t_0) & \boldsymbol{u}_{h}(\boldsymbol{\mu}_i, t_1)& \cdots & \boldsymbol{u}_{h}(\boldsymbol{\mu}_i, t_{N_{T_0}}) \\
\mid & \mid & & \mid
\end{array}\right] \in \mathbb{R}^{N \times (N_{T_0}+1)}, i = 1, \ldots, k,
\end{equation}
where $N_{T_0}+1$ is the number of the time steps sampled in $T_0, T_0 < T$. $k$ samples are taken from the parameter domain.

\subsection{Nonlinear dimensionality reduction via autoencoder}

\begin{figure}
	\centering
	\includegraphics[width=0.65\linewidth]{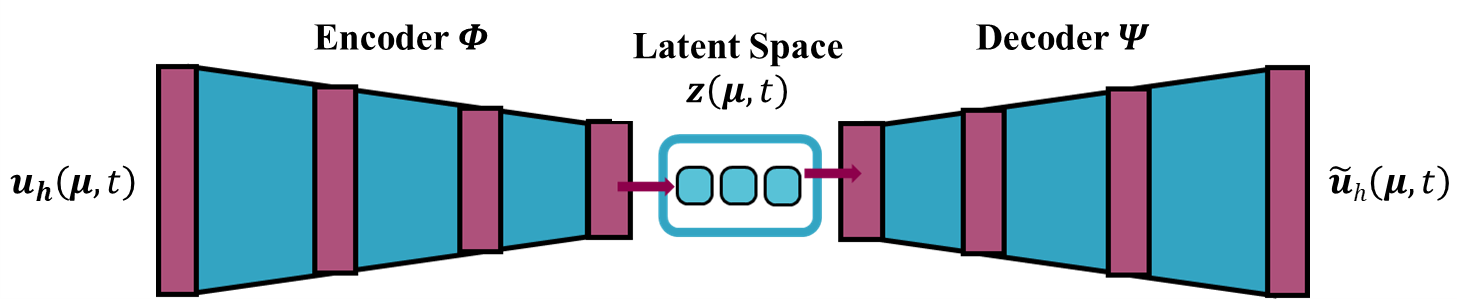}
	\caption{AE: AE generates a latent space with a nonlinear mapping.}
	\label{fig:CAE}
\end{figure}

Different from MOR methods based on linear projection, an autoencoder network, consisting of an encoder and a decoder, constructs a nonlinear mapping between the physical and the latent space. Autoencoder is an unsupervised machine learning tool with the loss being the difference between the input $\boldsymbol{u}_h(\boldsymbol{\mu}, t)$ and the reconstructed state $\tilde{\boldsymbol{u}}_h(\boldsymbol{\mu}, t)$. Spatial information is compressed into the latent space by the encoder and the latent variables are mapped back to the original physical space by the decoder. \Cref{fig:CAE} describes the structure of an autoencoder.

Neural networks such as FFNN, LSTM, SINDy and transformer~\cite{morMauLB21, FrHC22, morFreFM23, morConGFetal23, morGenZ22, morSolVGetal24} have been applied in the latent space to approximate the dynamics of the latent state. In this work,  we propose to use DMD as an effective alternative to capture the latent space dynamics in a relatively lightweight mode. 

\subsection{Kernel DMD}
DMD methods \cite{morSch10, morTuRLetal14} focus on predicting the time-evolution of the full state of the FOM in \cref{eq:ode}. In our case, KDMD \cite{morWilRK15} is applied for dynamics prediction in the latent space generated by the encoder. Assume that the dynamics evolution in the latent space is given by $\boldsymbol{z}(\boldsymbol{\mu}_i, t_{j+1}) = \boldsymbol{F}(\boldsymbol{z}(\boldsymbol{\mu}_i, t_j))$ at any given sample of $\boldsymbol{\mu}_i$, where $\boldsymbol{z} \in \mathbb{R}^n$, $n$ is the dimension of the latent space and $n \ll N$. $\boldsymbol{F}(\cdot)$ is the evolution function of $\boldsymbol{z}$. Based on the Koopman theory, $\boldsymbol{F}(\cdot)$ is assumed to be well approximated by the Koopman operator $\mathcal{K}$, i.e., 
\begin{equation}
\label{eq:Koopman_theory_1}
\boldsymbol{F}(\boldsymbol{z}(\boldsymbol{\mu},t)) \approx  \mathcal{K} \boldsymbol{z}(\boldsymbol{\mu},t).
\end{equation} 

In EDMD \cite{morWilKR15}, the infinite dimensional Koopman operator $\mathcal{K}$ is approximated by a finite dimensional matrix representation $\boldsymbol{K}$, which can be computed via $\boldsymbol{K} \triangleq \mathbf{\Psi}_{0}^{\dag} \mathbf{\Psi}_{1}$. The matrices $\mathbf{\Psi}_{0} \, , \mathbf{\Psi}_{1} \, \in \mathbb{R}^{N_{T_0} \times M}$ are defined as follows:

\begin{equation}
\mathbf{\Psi}_{0} \triangleq \left[\begin{array}{ccc}
\psi_{1}\left(\boldsymbol{z}_{0}\right) & \cdots & \psi_{M}\left(\boldsymbol{z}_{0}\right) \\
\psi_{1}\left(\boldsymbol{z}_{1}\right) & \cdots & \psi_{M}\left(\boldsymbol{z}_{1}\right) \\
\vdots & & \vdots \\
\psi_{1}\left(\boldsymbol{z}_{N_{T_0}-1}\right) & \cdots & \psi_{M}\left(\boldsymbol{z}_{N_{T_0}-1}\right)
\end{array}\right], \quad
\mathbf{\Psi}_{1} \triangleq \left[\begin{array}{ccc}
\psi_{1}\left(\boldsymbol{z}_{1}\right) & \cdots & \psi_{M}\left(\boldsymbol{z}_{1}\right) \\
\psi_{1}\left(\boldsymbol{z}_{2}\right) & \cdots & \psi_{M}\left(\boldsymbol{z}_{2}\right) \\
\vdots & & \vdots \\
\psi_{1}\left(\boldsymbol{z}_{N_{T_0}}\right) & \cdots & \psi_{M}\left(\boldsymbol{z}_{N_{T_0}}\right)
\end{array}\right],
\end{equation}
where $\psi_i( \cdot ), \, i = 1, \ldots, M$ are the $i$ different observables of $\boldsymbol{z}(\boldsymbol{\mu}, t)$. Assume that $\boldsymbol{z}(\boldsymbol{\mu}, t)$ can be represented by $N_{\ell}$ Koopman eigenfunctions $\phi_{\ell}(\cdot), \, \ell = 1, \ldots, N_{\ell}$, i.e., $\boldsymbol{z}(\boldsymbol{\mu}, t) = \sum_{\ell = 1}^{N_{\ell}} \phi_{\ell}(\boldsymbol{z}) \boldsymbol{v}_{\ell}$, where $ \boldsymbol{v}_{\ell}$ are the coefficients referred to as the Koopman modes. Based on the definition of eigenvalues and eigenfunctions, $\mathcal{K} \phi_{\ell}(\boldsymbol{z})= \lambda_{\ell} \phi_{\ell}(\boldsymbol{z}), \ell = 1,..., N_{\ell}$, , the evolution of $\boldsymbol{z}(\boldsymbol{\mu}, t)$ can be written as:

\begin{equation}
\label{eq:koopman_pred}
\boldsymbol{F}(\boldsymbol{z}) \approx \mathcal{K} \boldsymbol{z} = \sum_{\ell = 1}^{N_{\ell}}  \boldsymbol{v}_{\ell} \mathcal{K} \phi_{\ell}(\boldsymbol{z}) = \sum_{\ell = 1}^{N_{\ell}} \boldsymbol{v}_{\ell} \lambda_{\ell} \phi_{\ell}(\boldsymbol{z}).
\end{equation}

The tuple $(\lambda_{\ell}, \phi_{\ell}(\boldsymbol{z}), \mathbf{v}_{\ell})$, necessary for predicting the final solution, is derived from the eigenvalues and eigenvectors of $\boldsymbol{K}$. The eigenfunctions $\phi_{\ell}(\boldsymbol{z}) $ are computed by $\phi_{\ell}(\boldsymbol{z}) = \boldsymbol{\psi}(\boldsymbol{z}) \boldsymbol{w}_{\ell} $, where $\boldsymbol{w}_{\ell}$ is the $\ell$-th right eigenvector of the $\boldsymbol{K}$ and $\boldsymbol{\psi}(\boldsymbol{z})= [\psi_{1}(\boldsymbol{z})  \cdots   \psi_{M}(\boldsymbol{z})] \in \mathbb{R}^{1\times M}$ is a row-vector including $M$ observables given any $\boldsymbol{z}$. The eigenvalues of the Koopman operator $\lambda_{\ell}$ are approximated by the eigenvalues of $\boldsymbol{K}$. The Koopman modes $\mathbf{v}_{\ell}$ are determined via the left eigenvectors of $\boldsymbol{K}$. 

When $M \gg \max\{n, N_{T_0}\}$, the computation of the eigenvalues and the corresponding eigenvectors of $\boldsymbol{K} \in \mathbb{R}^{M \times M}$ requires large computational effort. This is the case for the problem we are considering, as $n \ll N$, and $N_{T_0}$ is usually small. For such a situation, an equivalent but less computationally expensive way to obtain these terms is proposed in \cite{morWilRK15} with the method KDMD.

After computing (truncated) singular value decomposition of $\mathbf{\Psi}_{0}$, i.e., $\mathbf{\Psi}_{0} = \mathbf{L} \mathbf{\Sigma} \boldsymbol{R}^T$, where $  \mathbf{L} \in \mathbb{R}^{N_{T_0} \times N_{T_0}}, \mathbf{\Sigma} \in \mathbb{R}^{N_{T_0} \times N_{T_0}}, \boldsymbol{R} \in \mathbb{R}^{M \times N_{T_0}}$, the eigenvalue problem $\lambda \boldsymbol{w} = \boldsymbol{K} \boldsymbol{w}$ can be reformulated via the transform  $\boldsymbol{w} = \boldsymbol{R} \hat{\boldsymbol{w}}$. This leads to a smaller-dimensional eigenvalue problem $\lambda \boldsymbol{R} \hat{\boldsymbol{w}} = \boldsymbol{R} \hat{\boldsymbol{K}} \hat{\boldsymbol{w}}$, where $\hat{\boldsymbol{K}} \triangleq \mathbf{\Sigma}^{-1}\mathbf{L}^T (\mathbf{\Psi}_1 \mathbf{\Psi}_0^T) \mathbf{L} \mathbf{\Sigma}^{-1} \in \mathbb{R}^{N_{T_0} \times N_{T_0}}$. The eigenvalue $\lambda_{\ell}$, the left eigenvector $\hat{\boldsymbol{\xi}}_{\ell}$ and the right eigenvector $\hat{\boldsymbol{w}}_{\ell}$ of $\hat{\boldsymbol{K}}$ can be used to compute the tuple $(\lambda_{\ell}, \phi_{\ell}(\boldsymbol{z}), \mathbf{v}_{\ell})$ more efficiently. The Koopman eigenfunctions $\phi_{\ell}(\boldsymbol{z})$ can be formulated as $\phi_{\ell}(\boldsymbol{z}) = \boldsymbol{\psi}(\boldsymbol{z}) \boldsymbol{R} \hat{\boldsymbol{w}}_{\ell} = \boldsymbol{\psi}(\boldsymbol{z}) \mathbf{\Psi}_0^T\mathbf{L} \mathbf{\Sigma}^{-1} \mathbf{\hat w}_{\ell}, {\ell}=1, \ldots, N_{\ell}$. The Koopman mode $\mathbf{v}_{\ell}$ is determined by the left eigenvector $\hat{\boldsymbol{\xi}}_{\ell}$ of $\hat{\boldsymbol{K}}$ via $\mathbf{v}_{\ell} =  (\boldsymbol{\hat \xi}_{\ell} \mathbf{\Sigma}^{-1} \mathbf{L}^{T} \boldsymbol{Z}_0^T)^T \in \mathbb R^n$, where $\boldsymbol{\hat \xi}_{\ell} \in \mathbb{R}^{1 \times N_{\ell}}$ is the ${\ell}$-th left eigenvector of the matrix $\mathbf{\hat K}$ satisfying $\boldsymbol{\hat \xi}_{\ell} \mathbf{\hat w}_{\ell} = 1$. $\boldsymbol{Z}_0$ is the data matrix in latent space, assembled as:

\begin{equation}
\boldsymbol{Z}_0= [\boldsymbol{z}(\boldsymbol{\mu}, t_0) \quad \boldsymbol{z}(\boldsymbol{\mu}, t_1) \; \cdots \; \boldsymbol{z}(\boldsymbol{\mu}, t_{N_{T_0} - 1})] \in \mathbb{R}^{n \times N_{T_0}}.
\end{equation}

In addition, $\mathbf{\Sigma}$ and $\mathbf{L}$ can also be obtained from the eigendecomposition of the matrix $\mathbf{\Psi}_0 \mathbf{\Psi}_0^T = \mathbf{L} \mathbf{\Sigma}^2 \mathbf{L}^T$. Note that each $i, j$-th entry of the matrix $\mathbf{\Psi}_0 \mathbf{\Psi}_0^T$ or $\mathbf{\Psi}_1 \mathbf{\Psi}_0^T$ can be computed via evaluating a kernel function $f(\boldsymbol{z}_{i-1}, \boldsymbol{z}_{j-1})$ or $f(\boldsymbol{z}_{i}, \boldsymbol{z}_{j-1})$, which can largely reduce the complexity of directly computing the matrix outer products, when $M$ is much larger than $n$.



The procedure of applying KDMD in the latent space is shown in \Cref{algorithm:kdmd}. The theory along with the derivation of KDMD for the FOM in \cref{eq:ode} are explained in detail in \cite{morWilRK15}.

\begin{algorithm}
	\begin{algorithmic}[1]
        \caption{Kernel DMD \cite{morWilRK15}} 
        \label{algorithm:kdmd}
        \State Compute $i,j$-th entry of $ \mathbf{\Psi}_0 \mathbf{\Psi}_0^T$ and $ \mathbf{\Psi}_1 \mathbf{\Psi}_0^T$ by a kernel function, i.e., $ (\mathbf{\Psi}_0 \mathbf{\Psi}_0^T)_{ij} = f(\boldsymbol{z}_{i-1}, \boldsymbol{z}_{j-1})$ and $ (\mathbf{\Psi}_1 \mathbf{\Psi}_0^T) =  f(\boldsymbol{z}_{i}, \boldsymbol{z}_{j-1})$, with $i,j = 1 , \ldots, m$.
        \State Compute the eigendecomposition of $\mathbf{\Psi}_0 \mathbf{\Psi}_0^T = \mathbf{L} \mathbf{\Sigma}^2 \mathbf{L}^T$.
        \State (optional) Choose the truncation rank $r$ that is smaller than the rank of $\mathbf{\Psi}_0 \mathbf{\Psi}_0^T$ to achieve a further reduction of the computation. Truncate the matrices $\mathbf{L}$, and $\mathbf{\Sigma}$ by keeping the first $r$ columns of $\mathbf{L}$ and first $r$ diagonal elements of $\mathbf{\Sigma}$ to obtain $\mathbf{L}_r$ and $\mathbf{\Sigma}_r$.
        \State Compute $\mathbf{\hat K} = (\mathbf{\Sigma}_{r}^{-1}\mathbf{L}_{r}^T) (\mathbf{\Psi}_1 \mathbf{\Psi}_0^T) (\mathbf{L}_{r} \mathbf{\Sigma}_{r}^{-1})$.
        \State Compute the eigendecomposition of $\mathbf{\hat K} \mathbf{\hat W} = \mathbf{\hat W} \mathbf{\hat \Lambda}$ with $\mathbf{\Lambda} = diag(\lambda_1,\ldots,\lambda_r), \mathbf{\hat W} = [\mathbf{\hat w}_1, \ldots, \mathbf{\hat w}_r]$.
        \State Compute the approximated Koopman eigenfuction $\phi_{\ell}(\boldsymbol{z}) = \boldsymbol{\psi}(\boldsymbol{z}) \mathbf{\Psi}_0^T\mathbf{L}_{r} \mathbf{\Sigma}_{r}^{-1} \mathbf{\hat w}_{\ell}$, where $\boldsymbol{\psi}(\boldsymbol{z}) \mathbf{\Psi}_0^T$ is obtained by evaluating the kernel function as $\boldsymbol{\psi}(\boldsymbol{z}) \mathbf{\Psi}_0^T = [f(\boldsymbol{z}, \boldsymbol{z}_0) \cdots f(\boldsymbol{z}, \boldsymbol{z}_{N_{T_0}-1})]$.  
        \State Set the Koopman modes as $\mathbf{v}_{\ell} =  (\boldsymbol{\hat \xi}_{\ell} \mathbf{\Sigma}_r^{-1} \mathbf{L}^{T}_{r} \boldsymbol{Z}_0^T)^T \in \mathbb R^n$, with ${\ell}=1, \ldots, r$.
        \State With the computed eigenvalues, eigenfunctions and Koopman modes $\lambda_{\ell}, \phi_{\ell}, \mathbf{v}_{\ell}$, the approximation of the evolution can be predicted via \cref{eq:koopman_pred} with $N_{\ell}=r$, i.e., $\boldsymbol{z}_{j+1} = \boldsymbol{F}(\boldsymbol{z}_{j})\approx \sum\limits_{{\ell}=1}^r \lambda_{\ell} \mathbf{v}_{\ell} \phi_{\ell} (\boldsymbol{z}_{j})$
	\end{algorithmic}
\end{algorithm}


\section{The Proposed Method}%
\label{sec:work}

In this section, we detail the building components of the newly proposed framework. Upon clarifying the functionalities of all components, the entire framework, referred to as the Data-Augmented Predictive Deep Neural Network, is assembled using these components. The proposed method merges KDMD into CAE to improve the time extrapolation performance of CAE-FFNN surrogate model.

\subsection{Building components}
\label{subsec:blocks}

The whole architecture is composed of several basic components. CAE constructs the latent space and FFNN builds the connection between the parameter-time space and the latent space. KDMD is responsible for the dynamics evolution in the latent space.

\subsubsection{CAE-FFNN}

\begin{figure}
	\centering
	\includegraphics[width=0.75\linewidth]{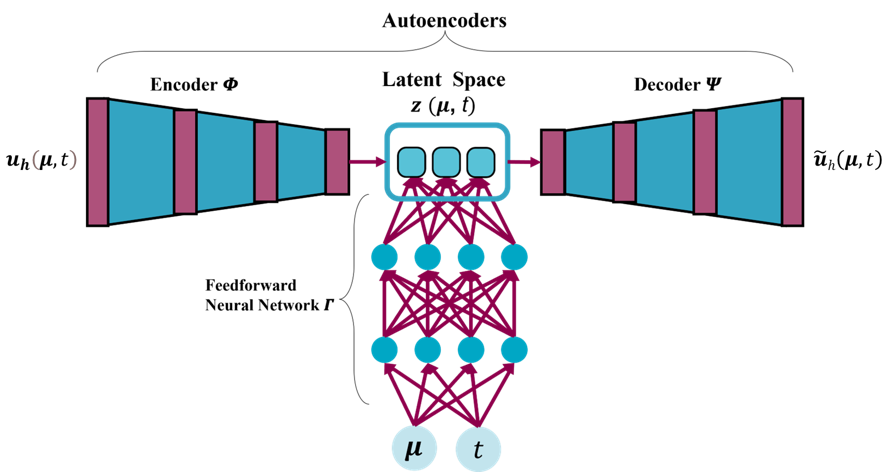}
	\caption{CAE-FFNN: CAE is responsible for the data compression and the data reconstruction. The FFNN connects the parameter-time space to the latent space.}
	\label{fig:CAE_FFNN}
\end{figure}

\Cref{fig:CAE_FFNN} shows the basic components CAE, FFNN and their combination CAE-FFNN. The encoder $\boldsymbol{\Phi}$ maps the original physical space to the latent space, while the decoder $\boldsymbol{\Psi}$ reconstructs the original physical dynamics. The FFNN $\boldsymbol{\Gamma}$ links the parameter-time to the latent space.

To be more detailed, the input of the autoencoder is the numerical solution of the FOM in \cref{eq:ode}, i.e., $\boldsymbol{u}_h(\boldsymbol{\mu}, t)$. Corresponding to that, the input of the FFNN is the parameter $\boldsymbol{\mu}$ and the time instant $t$. The optimization process of FFNN is controlled by the loss between the latent variable $\boldsymbol{z}(\boldsymbol{\mu}, t)$ i.e., the output of the encoder and the approximated latent variable that learned by the FFNN, i.e., the output of the FFNN. Finally, the loss of the full network can be formulated as:

\begin{equation}
\label{eq:loss_CAE_FFNN}
\mathcal{L}= \mathcal{L}_{CAE} + \alpha \mathcal{L}_{FFNN},
\end{equation}

\begin{equation}
\label{eq:loss_CAE}
\mathcal{L}_{CAE} = \frac{1}{k N_{T}} \sum_{i=1}^{k} \sum_{j=1}^{N_{T}} \left\|\boldsymbol{u}(\boldsymbol{\mu}_i, t_j)-\boldsymbol{\Psi}(\boldsymbol{\Phi}(\boldsymbol{u}(\boldsymbol{\mu}_i, t_j); \theta_{\boldsymbol{\Phi}});\theta_{\boldsymbol{\Psi}})\right\|_2^2,
\end{equation}

\begin{equation}
\label{eq:loss_FFNN}
\mathcal{L}_{FFNN} = \frac{1}{k N_{T}} \sum_{i=1}^{k} \sum_{j=1}^{N_{T}} \left\|\boldsymbol{\Phi}(\boldsymbol{u}(\boldsymbol{\mu}_i, t_j); \theta_{\boldsymbol{\Phi}})-\boldsymbol{\Gamma}(\boldsymbol{\mu}_i, t_j; \theta_{\boldsymbol{\Gamma}})\right\|_2^2,
\end{equation}

where $\mathcal{L}_{CAE}$ is the CAE reconstruction loss and $\mathcal{L}_{FFNN}$ is the FFNN loss. $ \theta_{\boldsymbol{\Phi}}, \theta_{\boldsymbol{\Psi}}$ and $\theta_{\boldsymbol{\Gamma}}$ are the neural network parameters that are learned during the training process. $\alpha$ is a hyperparameter that adjusts the contribution of the loss coming from the FFNN.

\subsubsection{CAE-KDMD}

\begin{figure}
	\centering
	\includegraphics[width=0.75\linewidth]{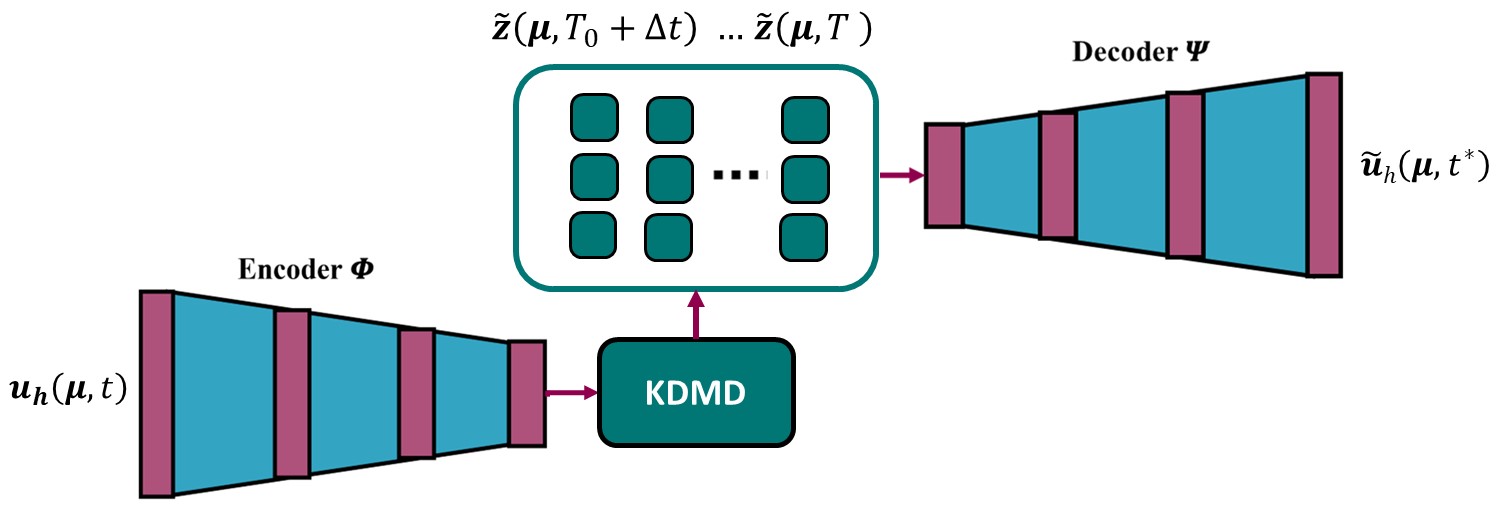}
	\caption{CAE-KDMD: In the latent space generated by CAE, the latent vector at any future time is extrapolated by KDMD and then decoded into the states in the physical space.}
	\label{fig:FFNN_DMD}
\end{figure}

\Cref{fig:FFNN_DMD} illustrates the combination of KDMD with CAE. After the latent space is learned by CAE, the prediction of the latent vectors at any future time is accomplished by KDMD.

First, a sequence of latent vectors $\boldsymbol{z}(\boldsymbol{\mu}, t)$ at training time period $t \in [0, T_0]$ are assembled into a snapshot matrix $\boldsymbol{Z}(\boldsymbol{\mu})$:
\begin{equation}
\label{eq:z_snapshots}
\boldsymbol{Z}(\boldsymbol{\mu})=\left[\begin{array}{cccc}
\mid & \mid & & \mid \\
\boldsymbol{z}(\boldsymbol{\mu}, t_1) & \boldsymbol{z}(\boldsymbol{\mu}, t_2)& \cdots & \boldsymbol{z}(\boldsymbol{\mu}, t_{N_{T_0}}) \\
\mid & \mid & & \mid
\end{array}\right] \in \mathbb{R}^{n \times N_{T_0}}.
\end{equation}

The output of KDMD is a new time sequence at the future time instants, i.e., $\tilde{\boldsymbol{z}}(\boldsymbol{\mu}, T_0 + j \Delta t), j = 1, 2, \ldots, N_T - N_{T_0}$, where $N_T \Delta t = T$. Through decoder, $\tilde{\boldsymbol{z}}(\boldsymbol{\mu}, T_0 + j \Delta t)$ can be mapped back to the physical space and the extrapolation $\tilde{\boldsymbol{u}}(\boldsymbol{\mu}, t^*)$ of the physical states in the time domain is finalized, where $t^* \in [T_0, T] $. To construct an efficient extrapolation map from $(\boldsymbol{\mu}, t)$ to the original physical solution $\boldsymbol{u}(\boldsymbol{\mu}, t)$ in both parameter and time domain, we need to combine the CAE-FFNN with the CAE-KDMD.

\subsection{Training networks with augmented data for prediction}%
\label{subsec:aug_data}

The proposed method is designed for fast and accurate prediction of the solution at any new testing parameter $\boldsymbol{\mu}^*$ and at any future time step $\hat t$ at the online phase. Here, $\boldsymbol{\mu}^*$ is not covered in the training parameters $\{\boldsymbol{\mu}_1, \boldsymbol{\mu}_2, \ldots, \boldsymbol{\mu}_k\}$ and $\hat t$ is in the time range of $[0, T], T > T_0$. Directly feeding $\hat t$ into the FFNN without any additional processing can result in inaccurate predictions for any time instants beyond the training time interval. To remedy this issue, a data-augmentation technique is integrated in our proposed framework.

\begin{figure}[ht]
	\centering
	\includegraphics[scale=0.475]{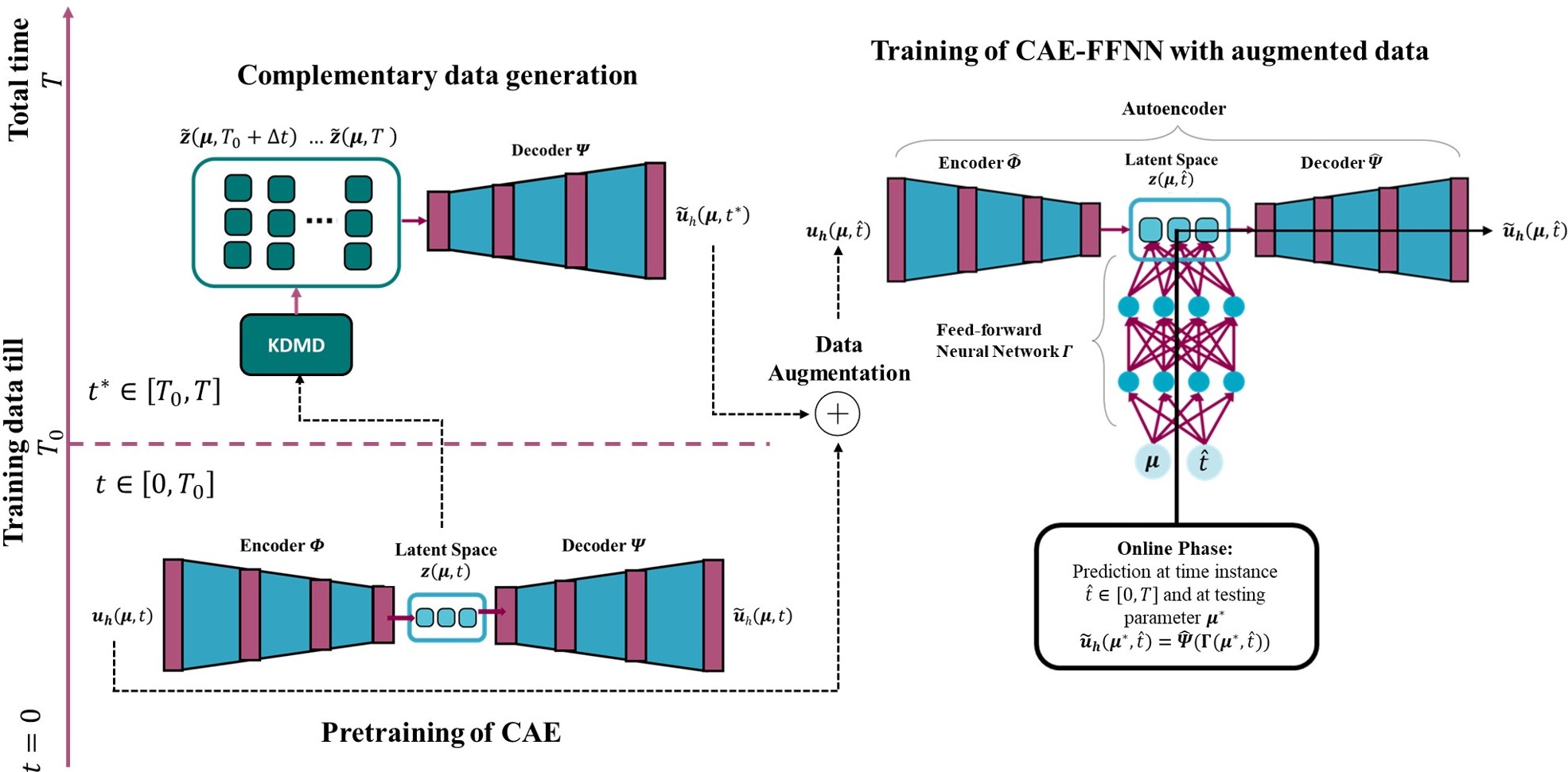}
	\caption{Data-augmented Predictive DNN flowchart: Lower-left part, pretraining of the CAE. Upper-left part, the generation of the complementary data. On the right side, train CAE-FFNN with the augmented data at the offline stage, and prediction at new parameter-time pairs with the trained FFNN-decoder in the online phase.}
	\label{fig:aug_data}
\end{figure}

The whole framework is assembled by three parts and proceeds in three corresponding steps. It is illustrated in \Cref{fig:aug_data}. The first step is the pretraining of CAE. The original training data is the solution snapshots at the training parameter samples $\{\boldsymbol{\mu}_1, \boldsymbol{\mu}_2, \ldots, \boldsymbol{\mu}_k\}$ in the training time interval $[0, T_0]$. CAE's training is controlled by the reconstruction loss, as shown in \cref{eq:loss_CAE}. This step includes initializing both the encoder and the decoder and generating the initial latent space data $\boldsymbol{z}(\boldsymbol{\mu}_i, t_j),  i=1,\ldots,k \, j = 1, \ldots, N_{T_0}$ in the time interval $[0, T_0]$. In the second step, the latent space data is assembled into the snapshot matrix in \cref{eq:z_snapshots}. KDMD uses the snapshot matrix $\boldsymbol{Z}(\boldsymbol{\mu})$ to predict the evolution of latent states, i.e., $\tilde{\boldsymbol{z}}(\boldsymbol{\mu}_i, t^*), t^* \in [T_0, T]$. The data predicted by KDMD is then fed into the decoder, yielding the reconstructed solution data $\tilde{\boldsymbol{u}}_h(\boldsymbol{\mu}_i, t^*), t^* \in [T_0, T]$ for every parameter sample in the training set. In the third step, the reconstructed data $\tilde{\boldsymbol{u}}_h(\boldsymbol{\mu}_i, t^*), t^* \in [T_0, T]$ is concatenated with the original data $\boldsymbol{u}_h(\boldsymbol{\mu}_i, t), t \in [0,T_0]$ to form the new training dataset $\boldsymbol{u}_h(\boldsymbol{\mu}_i, \hat t), \hat t \in [0,T]$ for CAE-FFNN. The CAE-FFNN is trained with this augmented data. During the online phase, given any parameter-time pair $(\boldsymbol{\mu}^*, \hat{t}), \hat t \in [0, T]$, FFNN predicts the latent state $\tilde{\boldsymbol{z}}(\boldsymbol{\mu}^*, \hat{t})$ and the newly trained decoder $\hat{\Psi}$ maps $\tilde{\boldsymbol{z}}(\boldsymbol{\mu}^*, \hat{t})$ to the final solution, i.e., $\tilde{\boldsymbol{u}}_h(\boldsymbol{\mu}^*, \hat{t})= \hat{\Psi}(\tilde{\boldsymbol{z}}(\boldsymbol{\mu}^*, \hat{t}))$. 

With the proposed method, extrapolation in the time domain is realized without relying on the training data ${\boldsymbol{u}}_h(\boldsymbol{\mu}_i, t^*), t^* \in [T_0, T]$, which is generated by expensive FOM simulations. For any given parameter-time input pair $(\boldsymbol{\mu}^*, \hat t), \hat t \in [0, T]$, the predicted solution $\tilde{\boldsymbol{u}}_h(\boldsymbol{\mu}^*, \hat t),  \hat t \in [0,T]$ can be directly obtained by the FFNN-decoder in one step, which is independent from sequentially predicting the solution at previous time steps. This framework avoids the $p$-step to $k$-step prediction in the online phase in the existing literature ~\cite{morMauLB21,  FrHC22, morFreFM23, morConGFetal23}. The training procedure and the testing phases are summarized in the \Cref{algorithm:DA-CAE-FFNN-off} and \Cref{algorithm:DA-CAE-FFNN-on}, respectively.

\begin{algorithm}
\hspace*{\algorithmicindent} \textbf{Input}:  The training parameters ${\boldsymbol{\mu}_1, \ldots, \boldsymbol{\mu}_k}$ and the time instants $t_1, \ldots, t_{N_{T_0}}$ in the training time interval $[0, T_0]$, the original training data $\boldsymbol{u}_h(\boldsymbol{\mu}_i, t_j), i = 1, \ldots, k, j = 1, \ldots, N_{T_0}$.

\hspace*{\algorithmicindent} \textbf{Output}: Neural network parameters of FFNN and those of the decoder, i.e., $\theta_{ \boldsymbol{\Gamma}} ,\theta_{ \boldsymbol{\hat \Psi}}$.
	\begin{algorithmic}[1]
        \caption{Framework of DAPredDNN - offline stage} 
        \label{algorithm:DA-CAE-FFNN-off}
        \State Train CAE with the original training data $\boldsymbol{u}_h(\boldsymbol{\mu}_i, t_j)$ to determine the neural network parameters $\theta_{\boldsymbol{\Phi}}, \theta_{\boldsymbol{\Psi}}$.
        \State Map the training data into the latent space by pretrained encoder, $\boldsymbol{z}(\boldsymbol{\mu}_i, t_j) = \boldsymbol{\Phi}(\boldsymbol{u}_h(\boldsymbol{\mu}_i, t_j))$ 
        \State For each $\boldsymbol{\mu}_i, i = 1, \ldots, k$, assemble the latent space data $\boldsymbol{z}(\boldsymbol{\mu}_i, t_j), t_j \in [0, T_0]$ into a snapshot matrix $\boldsymbol{Z}(\boldsymbol{\mu}_i)$ and feed it into KDMD in \Cref{algorithm:kdmd} to get $\tilde{\boldsymbol{z}}(\boldsymbol{\mu}_i, t^*), t^* \in [T_0, T]$
        \State Map the sequence of KDMD outputs  $\tilde{\boldsymbol{z}}(\boldsymbol{\mu}_i, t^*)$ back to the physical space by the pretrained decoder $\boldsymbol{\Psi}$, i.e., $\tilde{\boldsymbol{u}}(\boldsymbol{\mu}_i, t^*) = \boldsymbol{\Psi}(\tilde{\boldsymbol{z}}(\boldsymbol{\mu}_i, t^*))$.
        \State Concatenate the KDMD-extrapolated data $\tilde{\boldsymbol{u}}(\boldsymbol{\mu}_i, t^*), t^* \in [T_0, T]$ with the original training data $\boldsymbol{u}_h(\boldsymbol{\mu}_i, t), t \in [0,T_0]$ to form the augmented new training data. $\boldsymbol{u}(\boldsymbol{\mu}_i, \hat{t}) = \boldsymbol{u}_h(\boldsymbol{\mu}_i, t) \oplus \boldsymbol{u}(\boldsymbol{\mu}_i, t^*), \hat t \in [0, T]$.
        \State Train the CAE-FFNN with the augmented new training data corresponding to the parameter-time sample pairs $(\boldsymbol{\mu}_i, t_j), i = 1, \ldots, k$ and $t_j \in [0, T], j = 1, \ldots, N_T$ to determine the the neural network parameters of FFNN, i.e., $\theta_{\boldsymbol{\Gamma}}$ and update the parameters of CAE, i.e., $\theta_{\boldsymbol{\hat \Phi}}, \theta_{ \boldsymbol{\hat \Psi}}$.
	\end{algorithmic}
\end{algorithm}

\begin{algorithm}
\hspace*{\algorithmicindent} \textbf{Input}: Any testing parameters $ \boldsymbol{\mu}_1^*, \ldots,  \boldsymbol{\mu}_{k_{test}}^*$, any time instants $\hat t_1, \ldots, \hat t_{N_T}$ in the entire time interval $[0, T], T > T_0$ and the neural network parameters $\theta_{ \boldsymbol{\Gamma}} ,\theta_{ \boldsymbol{\hat \Psi}}$.

\hspace*{\algorithmicindent} \textbf{Output}: Predicted solution  $\tilde{\boldsymbol{u}}_h(\boldsymbol{\mu}^*, \hat t)$ at the inquired testing parameters and time instants.
	\begin{algorithmic}[1]
        \caption{Framework of DAPredDNN - online stage} 
        \label{algorithm:DA-CAE-FFNN-on}
        \State Load the neural network parameters $\theta_{ \boldsymbol{\Gamma}} ,\theta_{ \boldsymbol{\hat \Psi}}$ for the FFNN-decoder model.
        \State Arrange any inquired testing parameters $\boldsymbol{\mu}_i^*, i = 1,  \ldots, k_{test}$ and time instants $\hat t_j, j = 1,  \ldots, N_{T} $ into parameter-time pairs $(\boldsymbol{\mu}_i^*, \hat t_j)$ as inputs of the FFNN-decoder.
        \State Evaluate FFNN-decoder at those parameter-time pairs to obtain the corresponding predictions, $\tilde{\boldsymbol{u}}_h(\boldsymbol{\mu}^*_i, \hat t_j) = \boldsymbol{\hat \Psi}(\boldsymbol{\Gamma}(\boldsymbol{\mu}_i^*, \hat t_j))$.
	\end{algorithmic}
\end{algorithm}

\section{Numerical examples}
\label{sec:examples}

In this section, we test the performance of the proposed method on two numerical examples: a FitzHugh-Nagumo model and a model describing the flow past a cylinder. 

For assessing the model's accuracy, two error indicators are considered. The $\boldsymbol{\mu}$-$t$-dependent relative error $\boldsymbol{\epsilon}(\boldsymbol{\mu}, t) \in \mathbb{R}^N$ at any testing parameter sample $\boldsymbol{\mu}$ and at any time instant $t$ is defined as:
\begin{equation}
\label{eq:rel_err}
\boldsymbol{\epsilon}(\boldsymbol{\mu}, t_j) =   \frac{\lvert \boldsymbol{u}_h(\boldsymbol{\mu}, t_j) - \tilde{\boldsymbol{u}}_h((\boldsymbol{\mu}, t_j)) \rvert}{\sqrt{\sum_{j=1}^{N_T}\left\|\boldsymbol{u}_h(\boldsymbol{\mu}, t_j)\right\|^2}}.
\end{equation}

The maximum and mean values of the relative error over parameter, temporal and spatial domain are defined as:
\begin{equation}
\label{eq:rel_err_max}
\epsilon^{max} = \max_{i,j} \max_{k} \epsilon_k(\boldsymbol{\mu}_i, t_j), \quad \text{for all }\boldsymbol{\mu}_i \in \{\boldsymbol{\mu}^*_1, \ldots, \boldsymbol{\mu}^*_{k_{test}} \}, \, t_j \in [0, T], \, j = 1, \ldots, N_T,
\end{equation}

\begin{equation}
\label{eq:rel_err_mean}
\epsilon^{mean} = \frac{1}{k_{test} N_T N} \sum_{i,j,k}\epsilon_k(\boldsymbol{\mu}_i, t_j), \quad \text{for all }\boldsymbol{\mu}_i \in \{\boldsymbol{\mu}^*_1, \ldots, \boldsymbol{\mu}^*_{k_{test}} \}, \, t_j \in [0, T],  \, j = 1, \ldots, N_T,
\end{equation}
where $\epsilon_k$ is the $k$-th entry of $\boldsymbol{\epsilon}, \, k = 1, \ldots, N$. The other is the $\boldsymbol{\mu}$-dependent error indicator $\epsilon(\boldsymbol{\mu})$ defined as:

\begin{equation}
\label{eq:err_indicator}
\epsilon(\boldsymbol{\mu})=  \frac{\sqrt{\sum_{j=1}^{N_T}\left\|\boldsymbol{u}_h(\boldsymbol{\mu},t_j) - \tilde{\boldsymbol{u}}_h(\boldsymbol{\mu},t_j) \right\|^2}}{\sqrt{\sum_{j=1}^{N_T}\left\|\boldsymbol{u}_h(\boldsymbol{\mu},t_j)\right\|^2}}.
\end{equation}

The numerical examples are run on a personal computer equipped with 16 x 12th Gen Intel\textsuperscript\textregistered Core\textsuperscript\texttrademark i5-12600K, 31GB RAM, 64Bit CPU and NVIDA\textsuperscript\textregistered RTX\textsuperscript\texttrademark A4000 GPU. The code is implemented by Python 3.10 with Tensorflow 2.11.0 and Keras 2.11.0. In \Cref{subsubsec:FHNmodel}, we first test the model with the FitzHugh–Nagumo system. The data is simulated and obtained in the python code by solving the original system. In the second numerical example, the training data and reference testing data are collected by simulating the FOM in MATLAB \cite{Joh18matlab}.

\subsection{FitzHugh–Nagumo model}%
\label{subsubsec:FHNmodel}
The Fitz–Hugh Nagumo model is described by the following PDEs:

\begin{equation}
\begin{gathered}
\varepsilon v_{t}(x, \varepsilon, t)=\varepsilon^{2} v_{x x}(x, \varepsilon, t)+f(v(x, \varepsilon, t))-w(x, \varepsilon, t)+c \\
w_{t}(x,\varepsilon,  t)=b v(x, \varepsilon, t)-\gamma w(x, \varepsilon, t)+c.
\end{gathered}
\end{equation}
The boundary conditions are:
\begin{equation*}
\begin{array}{llr}
v(x, \varepsilon,0)=0, & w(x,\varepsilon, 0)=0, & x \in[0,L], \\
v_x(0,\varepsilon, t)=-i_o(t), & v_x(L, \varepsilon,t)=0, & t \geq 0,
\end{array}
\end{equation*}

The unknowns are $v(x,\varepsilon,t)$ representing the membrane potential, and $w(x,\varepsilon,t)$ representing the recovery of the potential. $f(v) = v(v-0.1)(1-v)$ is a cubic nonlinear term of the membrane potential. The parameter is $\boldsymbol{\mu} \coloneq \varepsilon \in [0.02, 0.04]$, while $L = 1.5$, $b = 0.5$, $c= 0.05$ and $\gamma = 2$. The input term $i_o(t) = 50000t^3e^{-15t}$ changes with time. 

The equation is discretized by the finite difference method, which results in a nonlinear time-dependent parametric system in the form of \Cref{eq:ode} with $N = 1024$. The numerical solution is $\boldsymbol{u}_h \coloneq [\boldsymbol{v}_h^T \; \boldsymbol{w}_h^T]^T \in \mathbb{R}^{N}$, where $ \boldsymbol{v}_h \in \mathbb{R}^{512}$ and $\boldsymbol{w}_h \in \mathbb{R}^{512}$.  The entire time intervalis $[0, T]$ with $T = 20$. The snapshots for training are only taken from the time interval $[0,T_0]$, i.e., $T_0 = 12$. $\Delta t$ is set to be $0.01$ leading to $N_T = 2000$.

In preparation of the dataset for the training stage, the snapshots of the numerical solution $\boldsymbol{v}_h$ and $\boldsymbol{w}_h$ under different parameters are gathered and aligned to form the training dataset.

The number of the training parameters is $k = 31$. These training parameters $\{\varepsilon_1, \ldots, \varepsilon_k\}$ are sampled equidistantly from $[0.01, 0.04]$. $k_{test} = 10$ testing parameters $\{\varepsilon^*_1, \ldots, \varepsilon^*_{k_{test}}\}$ are sampled randomly from $[0.01, 0.04]$ without overlapping the training parameters.

The architecture of CAE is shown in \Cref{tab:fhn_ae}. The encoder is constructed with five $1D$ convolutional layers and two dense layers. The number of the filters for each convolutional layer is $\{30, 25, 20, 15, 10\}$ with kernel size $ = 3$, stride $ = 1 $ and padding strategy is set as ``same". The pool sizes in MaxPooling1D layers are set as $2$. Two dense layers have the dimension of $\{ 32, 16 \}$. The activation functions in each layer are listed in \Cref{tab:fhn_ae}. The Swish activation function is defined as $swish(x) = x \cdot sigmoid(x)$. The dimension of the latent space is set to be $n = 2$. The decoder is constructed in a similar way and the MaxPooling1D layers are replaced by Upsampling1D layers.

\begin{table}[ht]
\begin{center}
\caption{FitzHugh-Nagumo model: The structure of CAE.}
\label{tab:fhn_ae}
\scalebox{0.8}{
\begin{tabular}{|ccc|}
\hline
\multicolumn{1}{|c|}{{\textbf{Layers}}}          & \multicolumn{1}{c|}{{\textbf{Output shape}}} & {\textbf{Activation function}} \\ \hline
\multicolumn{3}{|c|}{\textbf{Encoder}}                                                                            \\ \hline
\multicolumn{1}{|c|}{Input Layer}           & \multicolumn{1}{c|}{$1024 \times 1$}                   &                           \\ \hline
\multicolumn{1}{|c|}{Conv1D + MaxPooling1D} & \multicolumn{1}{c|}{$512 \times 30$}                   & Swish                     \\ \hline
\multicolumn{1}{|c|}{Conv1D + MaxPooling1D} & \multicolumn{1}{c|}{$256 \times 25$}                   & Swish                     \\ \hline
\multicolumn{1}{|c|}{Conv1D + MaxPooling1D} & \multicolumn{1}{c|}{$128 \times 20$}                   & Swish                     \\ \hline
\multicolumn{1}{|c|}{Conv1D + MaxPooling1D} & \multicolumn{1}{c|}{$64 \times 15$}                   & Swish                     \\ \hline
\multicolumn{1}{|c|}{Conv1D + MaxPooling1D} & \multicolumn{1}{c|}{$32 \times 10$}                   & Swish                     \\ \hline
\multicolumn{1}{|c|}{Flatten}               & \multicolumn{1}{c|}{$320$}                   &                           \\ \hline
\multicolumn{1}{|c|}{Dense}                 & \multicolumn{1}{c|}{$32$}                   & Swish                     \\ \hline
\multicolumn{1}{|c|}{Dense}                 & \multicolumn{1}{c|}{$16$}                   & Swish                     \\ \hline
\multicolumn{1}{|c|}{Dense (Output)}                 & \multicolumn{1}{c|}{$2$}                   & Linear                    \\ \hline
\multicolumn{3}{|c|}{\textbf{Decoder}}                                                                            \\ \hline
\multicolumn{1}{|c|}{Input}                 & \multicolumn{1}{c|}{$2$}                   &                           \\ \hline
\multicolumn{1}{|c|}{Dense}                 & \multicolumn{1}{c|}{$16$}                   & Swish                     \\ \hline
\multicolumn{1}{|c|}{Dense}                 & \multicolumn{1}{c|}{$32$}                   & Swish                     \\ \hline
\multicolumn{1}{|c|}{Reshape}               & \multicolumn{1}{c|}{$32 \times 1$}                   &                           \\ \hline
\multicolumn{1}{|c|}{Conv1D + Upsampling1D} & \multicolumn{1}{c|}{$64 \times 10$}                   & Swish                     \\ \hline
\multicolumn{1}{|c|}{Conv1D + Upsampling1D} & \multicolumn{1}{c|}{$128 \times 15$}                   & Swish                     \\ \hline
\multicolumn{1}{|c|}{Conv1D + Upsampling1D} & \multicolumn{1}{c|}{$256 \times 20$}                   & Swish                     \\ \hline
\multicolumn{1}{|c|}{Conv1D + Upsampling1D} & \multicolumn{1}{c|}{$512 \times 25$}                   & Swish                     \\ \hline
\multicolumn{1}{|c|}{Conv1D + Upsampling1D} & \multicolumn{1}{c|}{$1024 \times 30$}                   & Swish                     \\ \hline
\multicolumn{1}{|c|}{Conv1D (Output)}                & \multicolumn{1}{c|}{$1024 \times 1$}                   & Linear                    \\ \hline
\end{tabular}}
\end{center}
\end{table}

The structure of FFNN is displayed in \Cref{tab:fhn_ffnn}. The FFNN part is fully connected to $\boldsymbol{z}$ in the latent space. The dimension of five hidden layers in FFNN is set as $\{8, 16, 32, 64, 128\}$. The activation functions used in FFNN are listed in \Cref{tab:fhn_ffnn}.

\begin{table}[ht]
\begin{center}
\caption{FitzHugh-Nagumo model: The structure of FFNN.}
\label{tab:fhn_ffnn}
\scalebox{0.8}{
\begin{tabular}{|ccc|}
\hline
\multicolumn{1}{|c|}{{\textbf{Layers}}}          & \multicolumn{1}{c|}{{\textbf{Output shape}}} & {\textbf{Activation function}} \\ \hline
\multicolumn{3}{|c|}{\textbf{FFNN}}                                                                            \\ \hline
\multicolumn{1}{|c|}{Input Layer}           & \multicolumn{1}{c|}{$2$}                   &                           \\ \hline
\multicolumn{1}{|c|}{Dense}                 & \multicolumn{1}{c|}{$8$}                    & Swish                     \\ \hline
\multicolumn{1}{|c|}{Dense}                 & \multicolumn{1}{c|}{$16$}                   & Swish                     \\ \hline
\multicolumn{1}{|c|}{Dense}                 & \multicolumn{1}{c|}{$32$}                   & Swish                     \\ \hline
\multicolumn{1}{|c|}{Dense}                 & \multicolumn{1}{c|}{$64$}                   & Swish                     \\ \hline
\multicolumn{1}{|c|}{Dense}                 & \multicolumn{1}{c|}{$128$}                  & Swish                     \\ \hline
\multicolumn{1}{|c|}{Dense (Output)}        & \multicolumn{1}{c|}{$2$}                    & Linear                    \\ \hline
\end{tabular}}
\end{center}
\end{table}

Adam optimizer is applied with learning rate is $10^{-3}$. The CAE model is pretrained with $1000$ epochs using a batch size of $256$. After augmentation of the training data, the FFNN is pretrained with the new training data, where $10000$ epochs with a batch size of $256$ are applied. Then the CAE-FFNN model is trained after $20000$ epochs with a batch size of $256$. The hyperparameter $\alpha$ in the combined loss is set to be $0.1$ in \Cref{eq:loss_CAE_FFNN}. When training CAE-FFNN, we reduce the learning rate in the Adam optimizer by a factor of $1/2$ from $2\times 10^{-3}$ to a minimum value of $5 \times 10^{-4}$ whenever the loss stagnates. Gaussian kernel with shape parameter $=10$ is selected for KDMD in this case.

At the online phase shown in \Cref{fig:aug_data}, any testing parameters $\varepsilon^*_i$ in $\{\varepsilon^*_1, \ldots, \varepsilon^*_{k_{test}}\}$ and any time instant $\hat t_j$ in the time range $[0, T]$ are arranged as inputs $(\varepsilon^*_i, t_j), i = 1, \ldots, k_{test}, \, j = 1, \ldots, N_{T}$. The outputs are the predicted solution $v(x,\varepsilon^*_i,\hat t_j)$ and $w(x,\varepsilon^*_i,\hat t_j)$ at all testing parameters and all testing time instants in the entire time interval$[0, 20]$.

\begin{figure}
\begin{subfigure}{0.5\textwidth}
\includegraphics[width=0.94\linewidth]{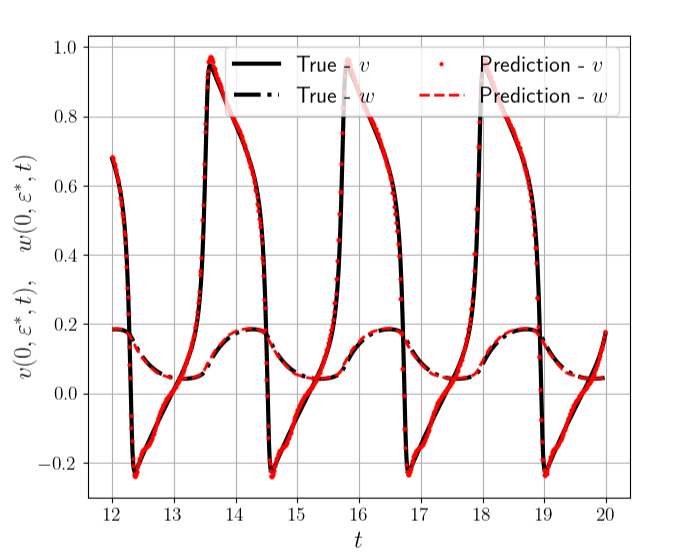}
\caption{}
\label{fig:FHN_model_sol_5}
\end{subfigure}
\begin{subfigure}{0.5\textwidth}
\includegraphics[width=0.99\linewidth]{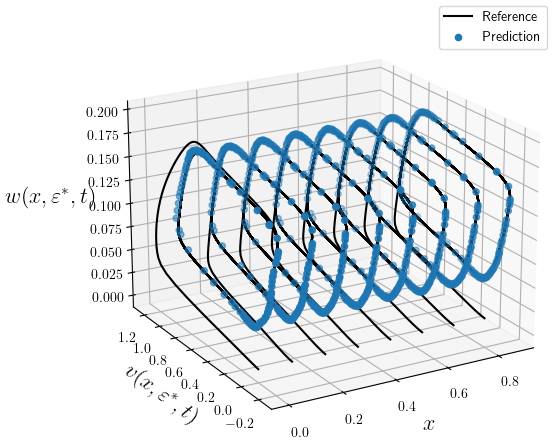}
\caption{}
\label{fig:FHN_model_sol_5_3d}
\end{subfigure}
\begin{subfigure}{0.5\textwidth}
\includegraphics[width=0.94\linewidth]{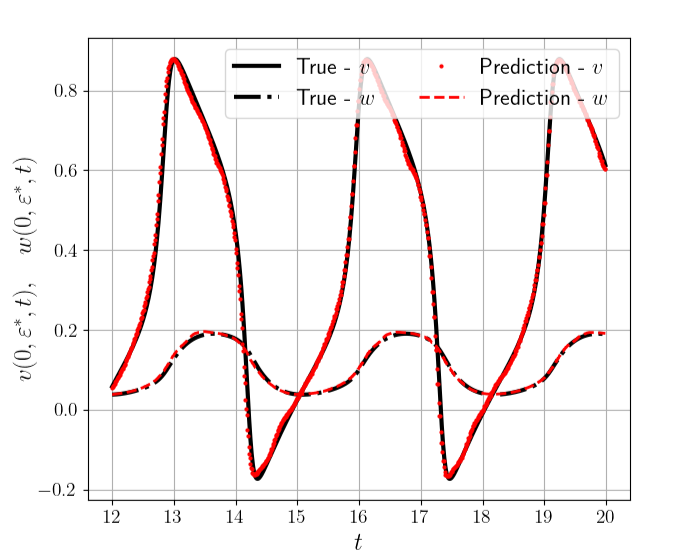} 
\caption{}
\label{fig:FHN_model_sol_9}
\end{subfigure}
\begin{subfigure}{0.5\textwidth}
\includegraphics[width=0.99\linewidth]{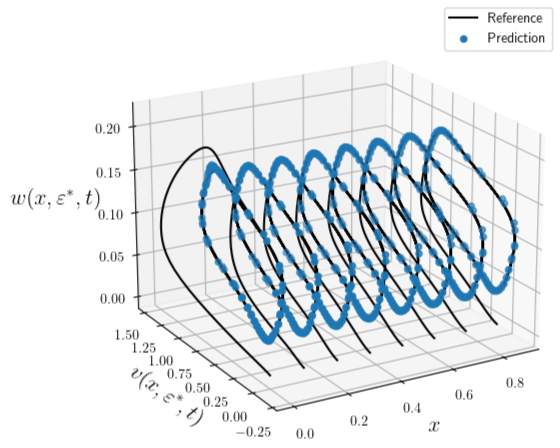}
\caption{}
\label{fig:FHN_model_sol_9_3d}
\end{subfigure}
\caption{FitzHugh-Nagumo model: the predicted solution and the reference solution. (a) The outputs $v(0,\varepsilon^*,t)$ and $w(0,\varepsilon^*,t)$ when $\varepsilon^* = 0.0151$. (b) Limit cycles of $v(x,\varepsilon^*,t)$ w.r.t. $w(x,\varepsilon^*,t)$ when $\varepsilon^* = 0.0151$. (c) The outputs $v(0,\varepsilon^*,t)$ and $w(0,\varepsilon^*,t)$ when $\varepsilon^* = 0.0352$. (d) Limit cycles of $v(x,\varepsilon^*,t)$ w.r.t. $w(x,\varepsilon^*,t)$ when $\varepsilon^* = 0.0352$.}
\label{fig:FHN_model_sol}
\end{figure}

\Cref{fig:FHN_model_sol} reports the predicted results at two randomly picked testing parameters in the time span $[T_0, T]$. \Cref{fig:FHN_model_sol_5} and \Cref{fig:FHN_model_sol_9} show the evolution of the two outputs $v(x = 0,\varepsilon^*,t)$ and $w(x = 0,\varepsilon^*, t)$ when $\varepsilon^* = 0.0151$ and $\varepsilon^* = 0.0352$, respectively. At the online stage, these output sequences at all the time instants $t_j, j = 1, \ldots, N_{T}$ in the time interval $[0, 20]s$ are predicted at one time by FFNN-decoder. The predicted solutions fit the references well. These results demonstrate that our proposed method is capable of predicting accurate results both in the time and the parameter space, even in the absence of training data in the extrapolation time span $[T_0, T]$. \Cref{fig:FHN_model_sol_5_3d} and \Cref{fig:FHN_model_sol_9_3d} are the phase-space diagrams showing $8$ cycles in the whole spatial domain.

\begin{figure}
\centering
\includegraphics[width=0.5\linewidth]{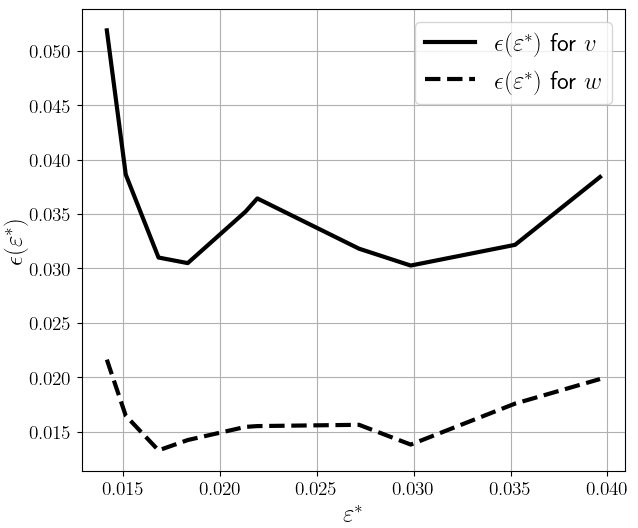}
\caption{FitzHugh-Nagumo model: the error indicators $\epsilon(\varepsilon^*)$ defined in \cref{eq:err_indicator} for the predicted solution $v(x, \varepsilon^*,t)$ and $w(x, \varepsilon^*,t)$ at all testing parameters, respectively.}
\label{fig:FHN_model_errind}
\end{figure}

\Cref{fig:FHN_model_errind} illustrates the values of the error indicator in \cref{eq:err_indicator} for the two solutions $v(x,\varepsilon^*,t)$ and $w(x,\varepsilon^*,t)$ at every testing parameter, respectively. The maximum and the mean errors defined in \cref{eq:rel_err_mean} are both listed in \Cref{tab:fhn_relerr}. The small errors indicate the reliability of the proposed method.

\begin{table}[ht]
\begin{center}
\caption{FitzHugh-Nagumo model: $\epsilon^{mean}$ and $\epsilon^{max}$ for $v$ and $w$, respectively.}
\label{tab:fhn_relerr}
\scalebox{0.8}{
\begin{tabular}{|c|c|c|}
\hline
  & $\epsilon^{mean}$ & $\epsilon^{max}$ \\ \hline
$v$ & $8.148 \times 10^{-4}$ & $1.970 \times 10^{-2}$            \\ \hline
$w$ & $4.479 \times 10^{-4}$ & $1.336 \times 10^{-2}$            \\ \hline
\end{tabular}}
\end{center}
\end{table}

\subsection{Incompressible flow past a cylinder}%
\label{subsubsec:cylinder_flow}

\begin{figure}
	\centering
	\includegraphics[width=0.625\linewidth]{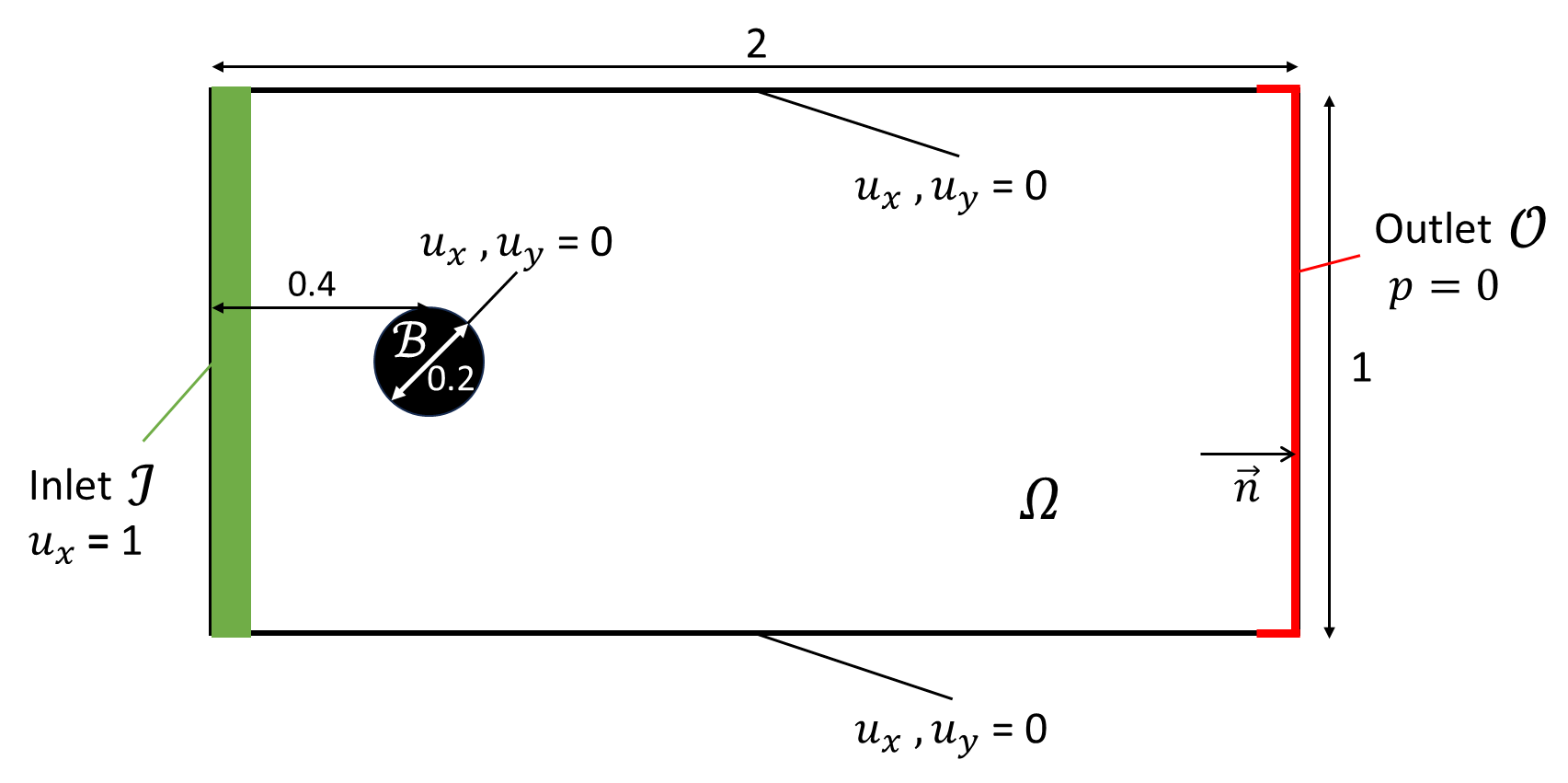}
	\caption{Flow around a cylinder: Geometry of a two-dimensional flow past a cylinder in a rectangular domain.}
	\label{fig:domain}
\end{figure}

This part shows the proposed framework applied to the model of the incompressible flow passing a cylinder. The evolution of the model is governed by 2D Navier-Stokes equations:

\begin{equation}
\begin{gathered}
\frac{\partial u_x}{\partial x}+\frac{\partial u_y}{\partial y}=0, \\
\frac{\partial u_x}{\partial t}+u_x \frac{\partial u_x}{\partial x}+u_y \frac{\partial u_x}{\partial y}=-\frac{1}{\rho} \frac{\partial p}{\partial x}+\nu\left(\frac{\partial^2 u_x}{\partial x^2}+\frac{\partial^2 u_x}{\partial y^2}\right), \\
\frac{\partial u_y}{\partial t}+u_x \frac{\partial u_y}{\partial x}+u_y \frac{\partial u_y}{\partial y}=-\frac{1}{\rho} \frac{\partial p}{\partial y}+\nu\left(\frac{\partial^2 u_y}{\partial x^2}+\frac{\partial^2 u_y}{\partial y^2}\right),
\end{gathered}
\end{equation}
where $u_x$ and $u_y$ are $x$ and $y$ components of the flow velocity $\boldsymbol{u}(x,y,\mu,t)$. The pressure is denoted as $p(x,y,\mu,t)$ and the density of the fluid is $\rho = 1.0$. $\nu$ is the kinematic viscosity with $\nu = 1/ Re$, where the Reynolds number $Re$ is chosen as the system parameter $\boldsymbol{\mu}$. The spatial domain $\Omega$, illustrated in \Cref{fig:domain}, is a rectangular domain $[0,2]$ $\times [0,1]$ with a cylindrical inclusion $\mathcal{B}$. The center of this circle locates at $(0.4, 0.4)$ with a radius of $0.2$. The inlet flow region $\mathcal{I}$ has a constant velocity, i.e., $u_x(\mathcal{I})= 1$, and the right boundary is the outlet flow region. The initial conditions are $\boldsymbol{u}(\Omega \setminus \mathcal{I}, t= 0) = \boldsymbol{0} $ and $p(x,y,t=0) = 0$. At the upper and lower wall $\boldsymbol{u}(x,y=0,t) = \boldsymbol{u}(x,y=1,t) = \boldsymbol{0} $. Neumann boundary condition is applied to the velocity at the outlet nodes as $\vv{\boldsymbol{n}} \cdot \mathbf{\nabla} \boldsymbol{u}(2,y,t) = \boldsymbol{0}$, where $\vv{\boldsymbol{n}}$ is the unit normal vector pointing to the boundary surface. The pressure is zero at the outlet region $p(2,y,t) = 0$.

Discretization of the governing equation in space is done by a finite difference scheme on a 2-dimensional spatial staggered grid with $\Delta x = \Delta y = 0.01$. For the time discretization, backward Euler differentiation scheme is applied with $\Delta t = 0.0015$. Given any $\boldsymbol{\mu}$, $t$, the discretized solution $\boldsymbol{u}_h \in \mathbb{R}^{N}, N = 18432$, is a vector of the magnitude of the velocity, i.e., $u = \sqrt{u_x^2 + u_y^2}$, at all the space grid points. Given any $\boldsymbol{\mu}$, $\boldsymbol{u}_h(\boldsymbol{\mu}, t_j), j = 1, \ldots, N_{T_0}$, is collected in the training data time interval $[4.5 , 8.1]$. $\boldsymbol{u}_h(\boldsymbol{\mu}^*_i, t_j), i = 1,  \ldots, k_{test}, \, j = 1, \ldots, N_T $ in the time interval $(8.1, 9]$ is unknown and to be predicted by the proposed method. The time interval $(8.1, 9]$ constitutes $25 \%$ extrapolation time compared to the entire time interval. The training set for $Re$ includes the following $k=24$ values: $\{$100, 125, 150, 175, 200, 225, 250, 275, 300, 325, 375, 400, 425, 450, 475, 500, 525, 600, 625, 650, 675, 700, 725, 750$\}$.


The detailed structure of CAE is listed in \Cref{tab:cyd_ae}. First, the CAE is pretrained alone and the latent space is initialized with $10$ latent variables, i.e., $n = 10$. The encoder is composed of five Conv2D layers with $\{8, 16, 16, 32, 32\}$ filters, respectively. Two fully connected network contains 2 hidden layers with $\{256, 64\}$ units. The decoder is constructed similarly. The kernel size is set as $3$ and the size of stride is set as $2$ for CAE. Adam optimizer is applied with learning rate $ 10^{-3}$. The batch size is $128$. Training CAE takes $5000$ epochs. Early stopping is added in the training if the loss fails to improve after $100$ epochs.

\begin{table}[ht]
\begin{center}
\caption{Flow past a cylinder model: The structure of CAE.}
\label{tab:cyd_ae}
\scalebox{0.8}{
\begin{tabular}{|ccc|}
\hline
\multicolumn{1}{|c|}{{\textbf{Layers}}}          & \multicolumn{1}{c|}{{\textbf{Output shape}}} & {\textbf{Activation function}} \\ \hline
\multicolumn{3}{|c|}{\textbf{Encoder}}                                                                            \\ \hline
\multicolumn{1}{|c|}{Input Layer}           & \multicolumn{1}{c|}{$96 \times 192 \times 1$}                   &                           \\ \hline
\multicolumn{1}{|c|}{Conv2D + MaxPooling2D} & \multicolumn{1}{c|}{$48 \times 96 \times 8$}                   & Swish                     \\ \hline
\multicolumn{1}{|c|}{Conv2D + MaxPooling2D} & \multicolumn{1}{c|}{$24 \times 48 \times 16$}                   & Swish                     \\ \hline
\multicolumn{1}{|c|}{Conv2D + MaxPooling2D} & \multicolumn{1}{c|}{$12 \times 24 \times 16$}                   & Swish                     \\ \hline
\multicolumn{1}{|c|}{Conv1D + MaxPooling2D} & \multicolumn{1}{c|}{$6 \times 12 \times 32$}                   & Swish                     \\ \hline
\multicolumn{1}{|c|}{Conv1D + MaxPooling2D} & \multicolumn{1}{c|}{$3 \times 6 \times 32$}                   & Swish                     \\ \hline
\multicolumn{1}{|c|}{Flatten}               & \multicolumn{1}{c|}{$576$}                   &                           \\ \hline
\multicolumn{1}{|c|}{Dense}                 & \multicolumn{1}{c|}{$256$}                   & Swish                     \\ \hline
\multicolumn{1}{|c|}{Dense}                 & \multicolumn{1}{c|}{$64$}                   & Swish                     \\ \hline
\multicolumn{1}{|c|}{Dense (Output)}                 & \multicolumn{1}{c|}{$10$}                   & Linear                    \\ \hline
\multicolumn{3}{|c|}{\textbf{Decoder}}                                                                            \\ \hline
\multicolumn{1}{|c|}{Input}                 & \multicolumn{1}{c|}{$10$}                   &                           \\ \hline
\multicolumn{1}{|c|}{Dense}                 & \multicolumn{1}{c|}{$64$}                   & Swish                     \\ \hline
\multicolumn{1}{|c|}{Dense}                 & \multicolumn{1}{c|}{$256$}                   & Swish                     \\ \hline
\multicolumn{1}{|c|}{Dense}                 & \multicolumn{1}{c|}{$576$}                   & Swish                     \\ \hline
\multicolumn{1}{|c|}{Reshape}               & \multicolumn{1}{c|}{$3 \times 6 \times 32$}                   &                           \\ \hline
\multicolumn{1}{|c|}{Conv2D + Upsampling2D} & \multicolumn{1}{c|}{$6 \times 12 \times 32$}                   & Swish                     \\ \hline
\multicolumn{1}{|c|}{Conv2D + Upsampling2D} & \multicolumn{1}{c|}{$12 \times 24 \times 32$}                   & Swish                     \\ \hline
\multicolumn{1}{|c|}{Conv2D + Upsampling2D} & \multicolumn{1}{c|}{$24 \times 48 \times 16$}                   & Swish                     \\ \hline
\multicolumn{1}{|c|}{Conv2D + Upsampling2D} & \multicolumn{1}{c|}{$48 \times 96 \times 16$}                   & Swish                     \\ \hline
\multicolumn{1}{|c|}{Conv2D + Upsampling2D} & \multicolumn{1}{c|}{$96 \times 192 \times 8$}                   & Swish                     \\ \hline
\multicolumn{1}{|c|}{Conv2D (Output)}                & \multicolumn{1}{c|}{$96 \times 192 \times 1$}                   & Linear                    \\ \hline
\end{tabular}}
\end{center}
\end{table}


The new training data is generated by augmenting the original training data in the time interval $[4.5, 8.1]$ with the KDMD-extrapolated data in the time interval $[8.1, 9]$. The structure of FFNN is illustrated in \Cref{tab:cyd_ffnn}. FFNN has four hidden dense layers with dimension $[8, 32, 64, 256]$ and one output layer connected to the latent space. Adam optimizer is employed with learning rate $10^{-3}$. Using batch size $128$, the pretraining of FFNN is processed with $5000$ epochs. Then CAE-FFNN is trained for $10000$ epochs with the same batch size and the same optimizer. The hyperparameter $\alpha$ in \Cref{eq:loss_CAE_FFNN} is set to be $1$. Besides, early stopping is enforced in the training of CAE-FFNN with a patience of $500$ epochs. The initial learning rate is set as $2\times 10^{-3}$ and is halved whenever the training loss fails to improve after $100$ epochs, until the minimum learning rate of $5\times 10^{-4}$ is reached.
 
\begin{table}[ht]
\begin{center}
\caption{Flow past a cylinder model: The structure of FFNN.}
\label{tab:cyd_ffnn}
\scalebox{0.8}{
\begin{tabular}{|ccc|}
\hline
\multicolumn{1}{|c|}{{\textbf{Layers}}}          & \multicolumn{1}{c|}{{\textbf{Output shape}}} & {\textbf{Activation function}} \\ \hline
\multicolumn{3}{|c|}{\textbf{FFNN}}                                                                            \\ \hline
\multicolumn{1}{|c|}{Input Layer}           & \multicolumn{1}{c|}{$2$}                   &                           \\ \hline
\multicolumn{1}{|c|}{Dense}                 & \multicolumn{1}{c|}{$8$}                    & Swish                     \\ \hline
\multicolumn{1}{|c|}{Dense}                 & \multicolumn{1}{c|}{$32$}                   & Swish                     \\ \hline
\multicolumn{1}{|c|}{Dense}                 & \multicolumn{1}{c|}{$64$}                   & Swish                     \\ \hline
\multicolumn{1}{|c|}{Dense}                 & \multicolumn{1}{c|}{$256$}                  & Swish                     \\ \hline
\multicolumn{1}{|c|}{Dense (Output)}        & \multicolumn{1}{c|}{$10$}                    & Linear                    \\ \hline
\end{tabular}}
\end{center}
\end{table}

During the online phase, we consider $6$ testing parameters $Re^* \in \{$180, 350, 550, 575, 710, 800$\}$, distinct from those used in training. The FFNN-decoder accurately predicts the corresponding solutions at any specified time instants within the entire time interval in one step.

\begin{figure}
\begin{subfigure}{1.0\textwidth}
	\centering
	\includegraphics[width=0.85\linewidth]{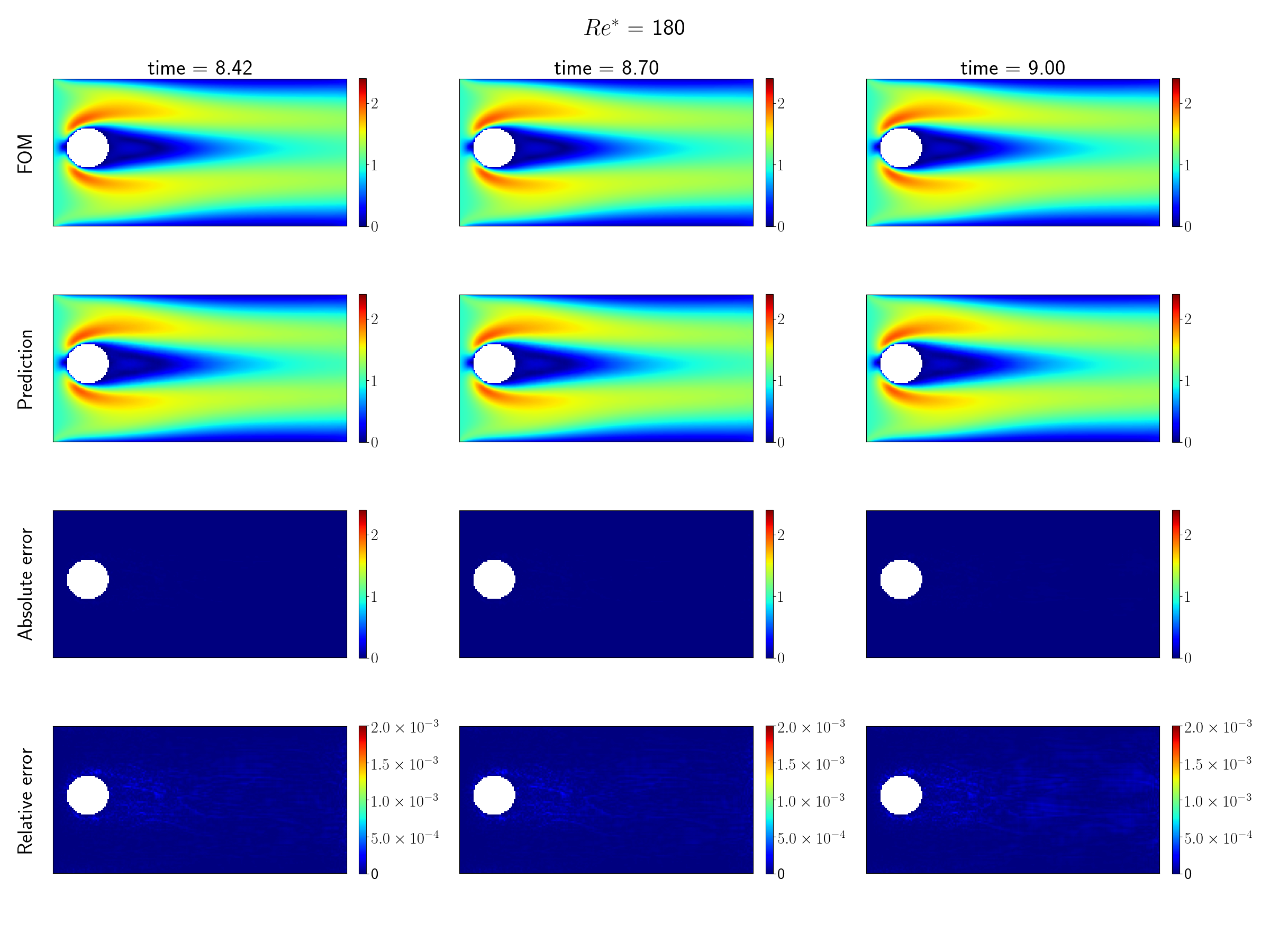}
	\caption{}
	\label{fig:cyflowresults0}
\end{subfigure}

\begin{subfigure}{1.0\textwidth}
	\centering
	\includegraphics[width=0.85\linewidth]{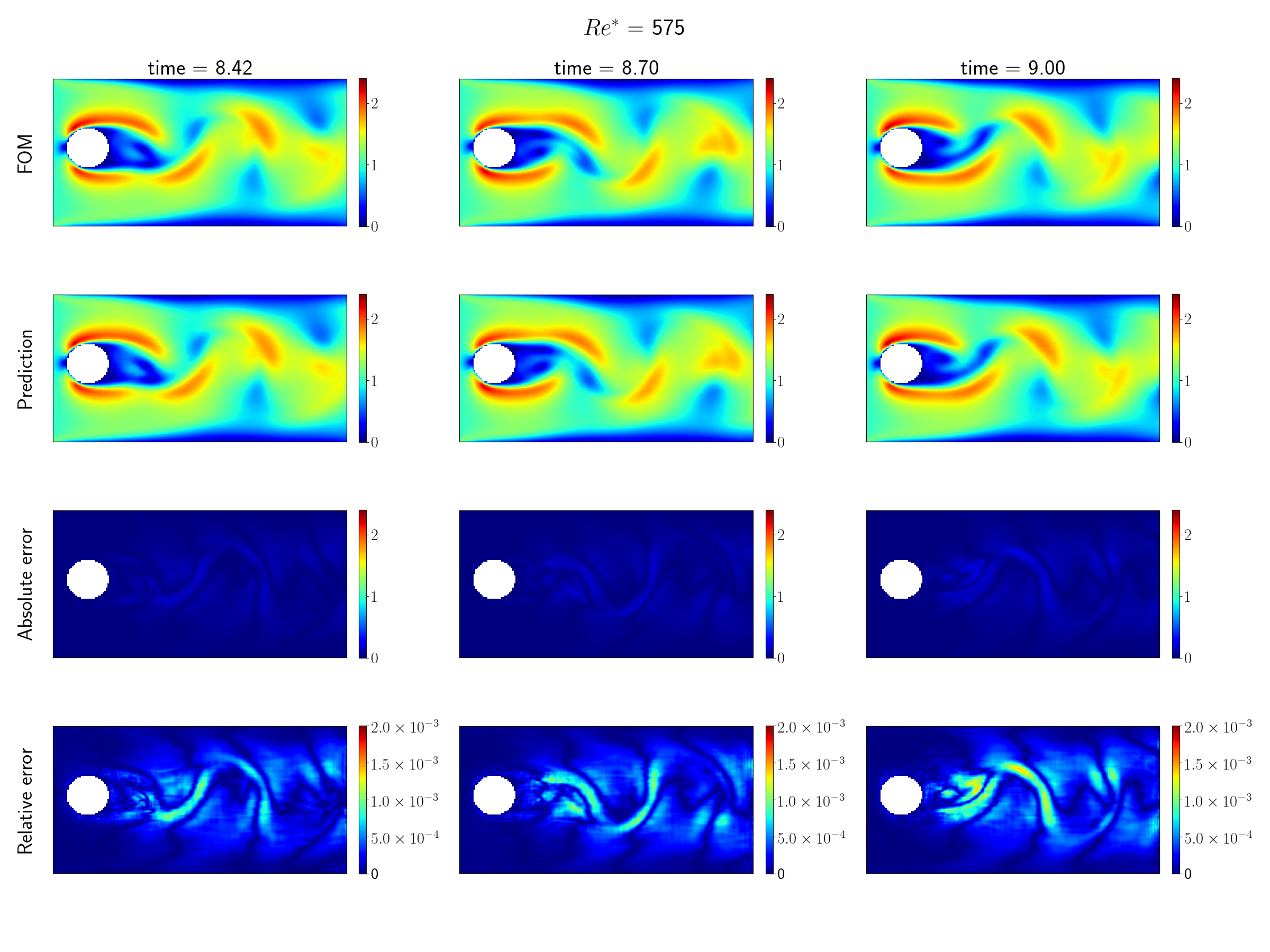}
	\caption{}
	\label{fig:cyflowresults1}
\end{subfigure}

\caption{Flow past a cylinder model: Magnitude of the velocity $\boldsymbol{u}$ at three time instants outside training time interval $[0, 8.1]$ and at two different testing parameters within the training parameter range $[100, 750]$, (a) $ Re^* = 180$, (b) $ Re^* = 575$.}
\label{fig:cyflowresults}
\end{figure}

\begin{figure}
	\centering
	\includegraphics[width=0.85\linewidth]{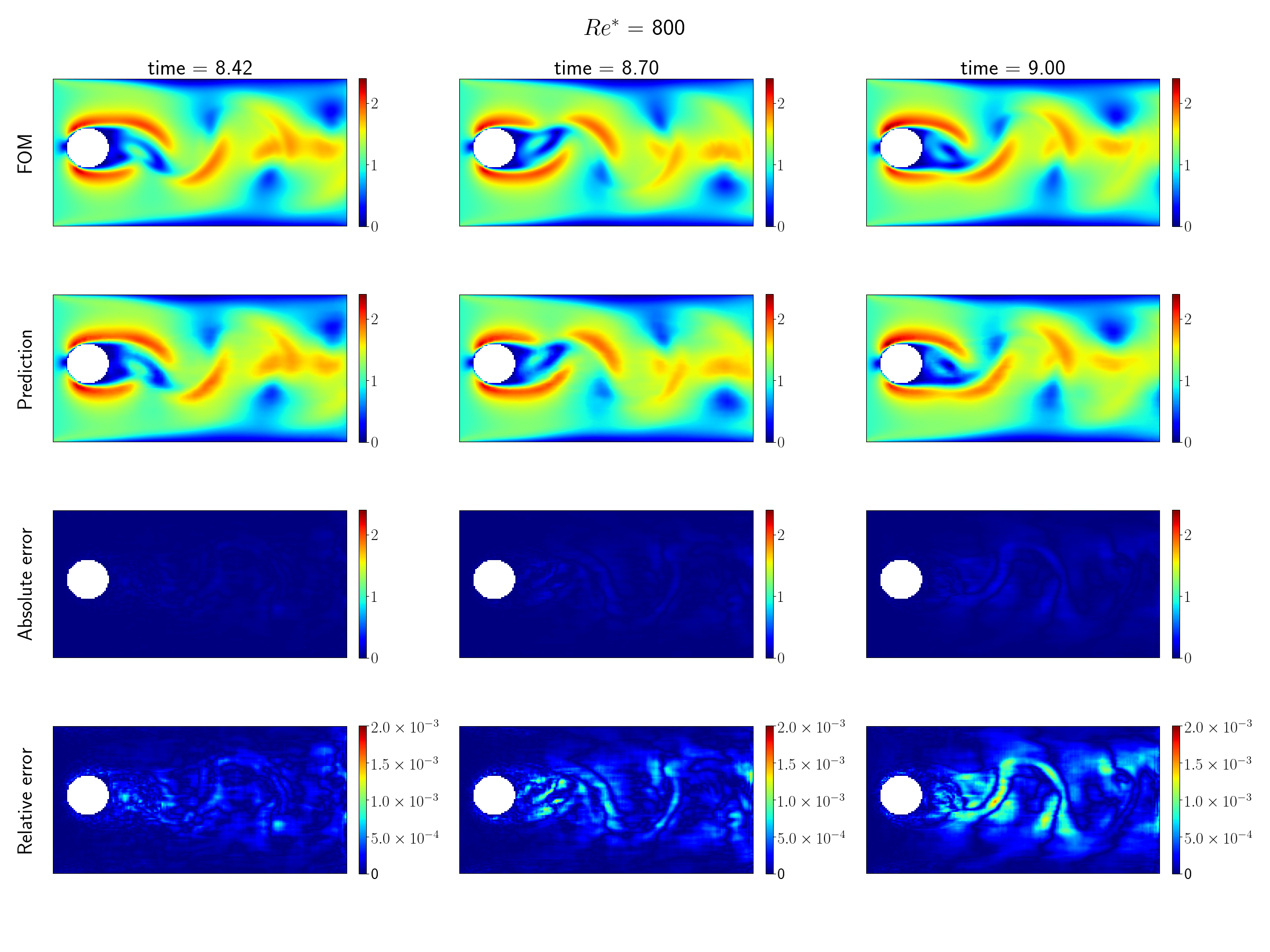}
	\caption{Flow past a cylinder model: Magnitude of the velocity $\boldsymbol{u}$ at three time instants outside training time interval $ [0, 8.1]$ and at $ Re^* = 800$ which is beyond the parameter training range $[100, 750]$.}
	\label{fig:cyflowresults2}
\end{figure}

\begin{figure}
\centering
\includegraphics[width=0.5\linewidth]{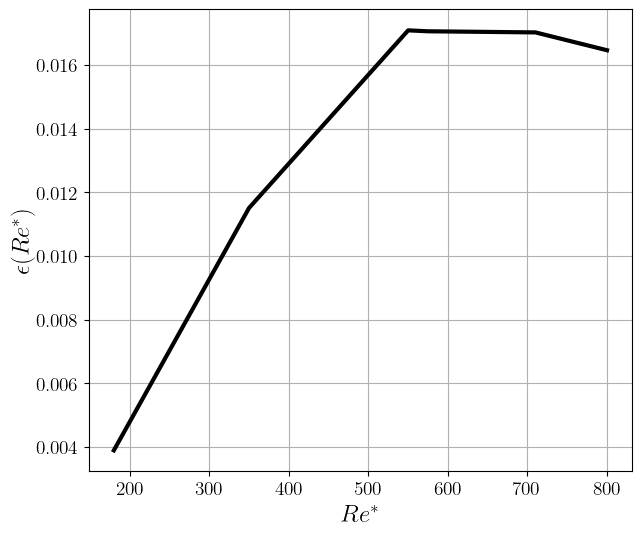}
\caption{Flow past a cylinder model: The error indicator $\epsilon(Re^*)$ in \cref{eq:err_indicator} at all testing parameters.}
\label{fig:Cy_errind}
\end{figure}

\Cref{fig:cyflowresults} and \Cref{fig:cyflowresults2} display the predicted results of three testing parameters $Re^*$ at three different time instants in the extrapolation time span, together with their reference results, as well as plots of the absolute errors and the relative errors. \Cref{fig:cyflowresults0} and \Cref{fig:cyflowresults1} show predictions at testing Reynolds number $Re^* =  180$ and $575$, which fall within the range of the training parameter  interval $[100, 750]$. From \Cref{fig:cyflowresults0}, we can observe that the steady laminar flow, with error at extrapolated time instants remaining significantly small. \Cref{fig:cyflowresults1} illustrates the vortex shedding, where the difference between the reference plots and the predicted solution is hardly detectable, too. \Cref{fig:cyflowresults2} presents the results at $Re^* = 800$, outside of the training range $[100, 750]$, showing the predicted vortex with a more complex evolution compared to the first two testing cases. The accuracy is still well preserved. The point-wise absolute error and the point-wise relative error in \cref{eq:rel_err} for each case are displayed in the last two rows of the \Cref{fig:cyflowresults} and \Cref{fig:cyflowresults2}, respectively. The values of the error indicator $\epsilon(Re^*)$ in \cref{eq:err_indicator} at all testing parameters are plotted in \Cref{fig:Cy_errind}. The maximum and the mean errors defined in\cref{eq:rel_err_max} and \cref{eq:rel_err_mean} for the testing time interval $[4.5 , 9.0]$ are listed in \Cref{tab:cy_relerr}.

\begin{table}[ht]
\begin{center}
\caption{Flow past a cylinder model: $\epsilon^{mean}$ and $\epsilon^{max}$.}
\label{tab:cy_relerr}
\scalebox{0.8}{
\begin{tabular}{|c|c|}
\hline
$\epsilon^{mean}$ & $\epsilon^{max}$ \\ \hline
$5.498 \times 10^{-5}$ & $1.888 \times 10^{-3}$            \\ \hline
\end{tabular}}
\end{center}
\end{table}

All these results demonstrate the method's capability to accurately predict solutions at time instants beyond the training range across all testing parameters, showing the reliability of the proposed method.
%
\section{Conclusion and outlook}
\label{sec:conclu}

This work presents a non-intrusive parametric MOR method employing the deep learning structures CAE-FFNN and KDMD. The framework is described in detail and validated with two numerical examples. These results indicate that the proposed method provides an accurate prediction in both the time and the parameter space.

The novelties of this work include: First, leveraging KDMD in the latent space enables the framework to predict dynamics beyond the limited time interval of the training data. Second, the proposed method enriches original data by augmenting the KDMD-decoder-extrapolated data during the offline stage. Training CAE-FFNN with this augmented data allows the network to capture future features. Thirdly, during the online phase, the model is efficient by only evaluating the FFNN-decoder without involving the KDMD. The FFNN-decoder directly correlates any time-parameter pairs with their corresponding extrapolated solutions, avoiding the time-marching behaviour that limits the efficiency and the accuracy of the online prediction.

In future research, we may introduce a discriminator into the training process to validate the quality of data generated by KDMD. This addition could enhance both the accuracy and stability of the framework while reducing the efforts required for the hyperparameter tuning.

\section*{Data and code availability}
\addcontentsline{toc}{section}{Data and code availability}
Data and code are available upon reasonable request and will be made public in Zenodo - \url{https://doi.org/10.5281/zenodo.13940888} upon publication of this manuscript.

\section*{Acknowledgments}%
\addcontentsline{toc}{section}{Acknowledgments}
This research is supported by the International Max Planck Research School for Advanced Methods in Process and Systems Engineering (IMPRS ProEng), Magdeburg, Germany.


\addcontentsline{toc}{section}{References}
\bibliographystyle{plainurl}
\bibliography{refs_new.bib}
  
\end{document}